\documentclass[10pt,twocolumn,letterpaper]{article}

\usepackage{iccv}              

\usepackage[accsupp]{axessibility}  

\usepackage{amsmath,amssymb,amsfonts}
\usepackage{graphicx}
\usepackage{textcomp}
\usepackage{xcolor, array, colortbl}

\usepackage[utf8]{inputenc}
\usepackage[T1]{fontenc}    
\usepackage{url}          
\usepackage{booktabs}      
\usepackage{amsfonts}      
\usepackage{nicefrac}    
\usepackage{microtype} 

\usepackage{bbding}
\usepackage{pifont}
\usepackage{amssymb}
 
\newcommand{\xmarkg}{\textcolor{gray}{\ding{55}}\xspace}%
\newcommand{\cmark}{\ding{51}}
\newcommand{\tabincell}[2]{}
\usepackage{bm}
\usepackage{comment}
\usepackage{enumerate} 
\usepackage{enumitem}
\usepackage{mathrsfs}
\usepackage{makeidx}      
\usepackage{graphicx}     
\usepackage{epsfig,latexsym}
\usepackage{psfrag}
\usepackage[ruled,vlined]{algorithm2e}
\definecolor{lightgray}{rgb}{.93,.93,.93}
\definecolor{deepred}{rgb}{0.698,0.133,0.133}
\definecolor{blue}{rgb}{0,0,1}

\usepackage{makecell}	
\usepackage{fancyhdr}
\usepackage{multirow}
\usepackage{rotating}
\usepackage{color}
\usepackage{mathtools}
\usepackage{amsmath}
\usepackage{tensor}

\newcommand{\mf}{\mathbf}

\newcommand{\mr}{\mathrm}

\definecolor{iccvblue}{rgb}{0.21,0.49,0.74}
\usepackage[pagebackref,breaklinks,colorlinks,allcolors=blue]{hyperref}

\def\paperID{4655} 
\def\confName{ICCV}
\def\confYear{2025}

\title{Hierarchical Visual Prompt Learning for Continual Video Instance Segmentation}

\author{
Jiahua Dong\textsuperscript{1}\footnotemark[1]~,
Hui Yin\textsuperscript{2}\footnotemark[1]~,
Wenqi Liang\textsuperscript{3},
Hanbin Zhao\textsuperscript{4},
Henghui Ding\textsuperscript{5}, 
Nicu Sebe\textsuperscript{3}, \\
Salman Khan\textsuperscript{1,6},
Fahad Shahbaz Khan\textsuperscript{1,7} \\
\textsuperscript{1}Mohamed bin Zayed University of Artificial Intelligence
~\textsuperscript{2}Hunan University ~\textsuperscript{3}University of Trento \\
\textsuperscript{4}Zhejiang University ~\textsuperscript{5}Fudan University
~\textsuperscript{6}Australian National University
~\textsuperscript{7}Link\"{o}ping University \\
{\tt\small \{dongjiahua1995, yinhui2000a, liangwenqi0123, henghui.ding\}@gmail.com, zhaohanbin@zju.edu.cn} \\  
{\tt\small niculae.sebe@unitn.it, \{salman.khan, fahad.khan\}@mbzuai.ac.ae}
}

\begin{document}
\maketitle

\renewcommand{\thefootnote}{\fnsymbol{footnote}}
\footnotetext[1]{Equal contributions. The corresponding authors are Dr. Jiahua Dong and Prof. Henghui Ding. This work was supported in part by the National Nature Science Foundation of China under Grant 62133005 and 62472104. }

\begin{abstract}
Video instance segmentation (VIS) has gained significant attention for its capability in tracking and segmenting object instances across video frames. However, most of the existing VIS approaches unrealistically assume that the categories of object instances remain fixed over time. Moreover, they experience catastrophic forgetting of old classes when required to continuously learn object instances belonging to new categories. To resolve these challenges, we develop a novel \underline{H}ierarchical \underline{V}isual \underline{P}rompt \underline{L}earning (HVPL) model that overcomes catastrophic forgetting of previous categories from both frame-level and video-level perspectives. Specifically, to mitigate forgetting at the frame level, we devise a task-specific frame prompt and an orthogonal gradient correction (OGC) module. The OGC module helps the frame prompt encode task-specific global instance information for new classes in each individual frame by projecting its gradients onto the orthogonal feature space of old classes. 
Furthermore, to address forgetting at the video level, we design a task-specific video prompt and a video context decoder. This decoder first embeds structural inter-class relationships across frames into the frame prompt features, and then propagates task-specific global video contexts from the frame prompt features to the video prompt. 
Through rigorous comparisons, our HVPL model proves to be more effective than baseline approaches. 
The code is available at \url{https://github.com/JiahuaDong/HVPL}.

\end{abstract}

\section{Introduction}
Video instance segmentation (VIS) \cite{10376837, 10.1007/978-3-031-72667-5_6, Huang_2024_CVPR, huang2022minvis, SegSurvey} is a sophisticated vision task that requires simultaneously detecting, segmenting, and tracking individual object instances across consecutive frames. Unlike traditional object detection \cite{NIPS2015_14bfa6bb} or segmentation \cite{9710959}, which predicts within individual frames, VIS seeks to detect and segment each unique object instance across an entire video sequence, maintaining consistent tracking over time \cite{Ying_2023_ICCV, 10.1007/978-3-031-19815-1_32}. This requires distinguishing individual objects as they move, change appearance, or undergo occlusions, making VIS a particularly challenging research field \cite{maskfreevis, CERON2022102569}.
In recent years, VIS has become essential in various applications such as autonomous driving \cite{9575445}, robotics \cite{9093335}, and augmented reality \cite{10024449}.

\begin{figure}[t]
\centering
\includegraphics[trim = 5mm 11mm 5mm 11mm, clip, width=236pt, height=160pt]
{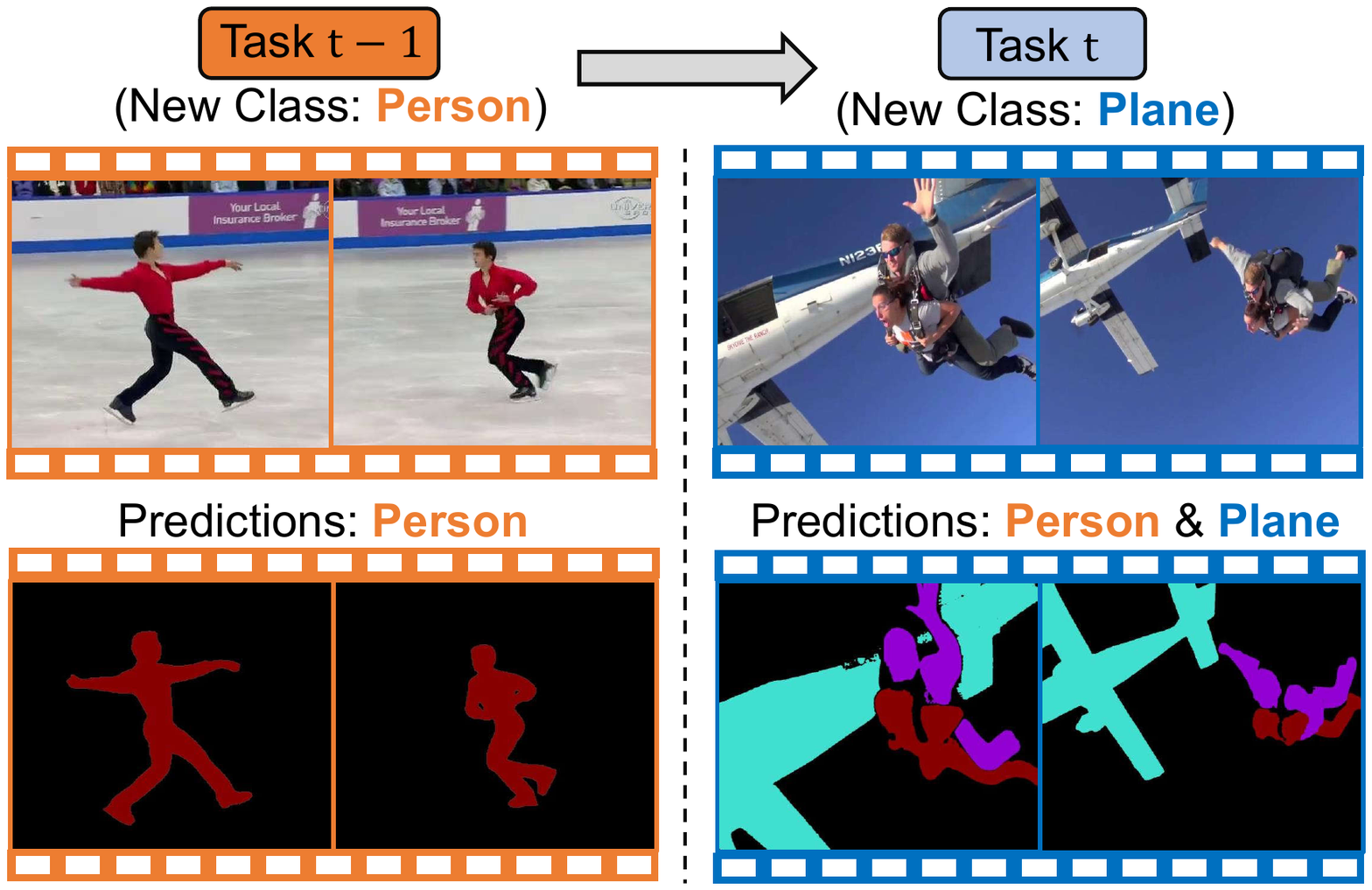}
\vspace{-20pt}
\caption{Demonstration of continual video instance segmentation (CVIS) for learning object instances belonging to new classes consecutively. At the $t$-th task, CVIS seeks to learn new categories (\emph{e.g.}, \texttt{plane}) while alleviating forgetting on old ones (\emph{e.g.}, \texttt{person}). }
\label{fig: motivation}
\end{figure}

Generally, most existing VIS methods \cite{fang2024unified, VITA, 9578755, cheng2021mask2forme} are designed for static application scenarios, where the categories of object instances are predefined and remain constant over time. However, real-world scenarios are typically dynamic, with object instances of new classes arriving consecutively in an online manner. To tackle this limitation, traditional VIS models \cite{ke2021prototypical, Li_2023_ICCV, 9577716} often require retaining all video data of old categories to retrain the entire model. Nevertheless, as the number of old categories grows continuously, this leads to high computational costs and memory requirements, rendering existing VIS methods \cite{huang2022minvis, 10.1007/978-3-031-72667-5_6} impractical. If the above VIS methods \cite{10204984, cheng2023putting, Athar_Mahadevan20ECCV} lack memory to retain video data for old classes, they may exhibit substantial accuracy drops for previously learned categories, known as the catastrophic forgetting phenomenon \cite{Rebuffi_2017_CVPR, wang2023hierarchical, dong2022federated_FCIL}.

To handle the aforementioned practical scenarios, we propose a new problem named \textbf{\textit{Continual Video Instance Segmentation}} (CVIS) in this paper. Unlike conventional VIS, which assumes a fixed set of object classes, the proposed CVIS problem seeks to incrementally learn new object categories while alleviating forgetting of old ones, as demonstrated in Fig.~\ref{fig: motivation}. Considering the practical application of CVIS in real-world scenarios, we do not allocate any memory to retain training data of old classes, nor do we have prior knowledge about the distribution of new classes. An intuitive approach to tackle the CVIS problem is to simply combine continual learning (CL) \cite{9710595, zhao2021memory, 10555533271443327290} with video instance segmentation (VIS) \cite{9577282, 9878483}. However, this strategy primarily aims to address catastrophic forgetting at the frame level, and thus struggles to explore the global video context of new classes to alleviate forgetting at the video level. In particular, the lack of global video context can significantly hinder the effectiveness of this simple strategy in overcoming catastrophic forgetting from the video-level perspective.

To resolve the challenges posed by the CVIS problem, we develop a novel \underline{H}ierarchical \underline{V}isual \underline{P}rompt \underline{L}earning (HVPL) model, which effectively mitigates catastrophic forgetting of old classes from both frame-level and video-level perspectives. \textbf{On one hand}, we propose a task-specific frame prompt and an orthogonal gradient correction module to alleviate forgetting at the frame level. After using Mask2Former \cite{9878483} as a frame-level detector to extract latent features frame by frame for a given video, the frame prompt can encode task-specific global instance information for new classes in each individual frame. Then the orthogonal gradient correction module projects the gradients used to update the frame prompt onto the orthogonal feature space of old classes for retaining previous knowledge. 
\textbf{On the other hand}, to address the forgetting of old categories from the video-level perspectives, we devise a task-specific video prompt and a video context decoder. After this decoder embeds structural inter-class relationships across frames into the frame prompt features, it captures task-specific global video contexts from the frame prompt features and propagates relevant information to the video prompt for forgetting compensation. Experiments on VIS datasets verify the competitive improvements of our HVPL model over other baselines. We summarize the primary contributions of this paper below: 
\begin{itemize}
\item We propose a new practical problem called Continual Video Instance Segmentation (CVIS). To address the CVIS problem, we develop a novel Hierarchical Visual Prompt Learning (HVPL) model, which effectively addresses forgetting from both frame-level and video-level perspectives.

\item We design a task-specific frame prompt and an orthogonal gradient correction module to alleviate forgetting at the frame level by projecting the gradients used to update frame prompt onto orthogonal feature space of old classes.

\item We propose a task-specific video prompt and a video context decoder to mitigate forgetting from the video-level perspective by capturing task-specific global video contexts for new classes from the frame prompt features. 
\end{itemize}

\section{Related Work}
\textbf{Video Instance Segmentation} (VIS) \cite{thawakar2024videoinstances, 10204941, Wang_2024_CVPR, MOSE_ding} combines object detection, instance segmentation, and temporal tracking within videos in computer vision. It is typically divided into two main categories: online and offline methods. Unlike \cite{9879517, ke2021prototypical, 9710024}, online VIS methods leverage query-based detectors \cite{10.1007/978-3-030-58452-8_13, 9578639}. 
\cite{huang2022minvis} applies query-based detectors independently to video frames and tracks object instances using bipartite matching. To address the semantic ambiguity, \cite{10.1007/978-3-031-19815-1_34, Ying_2023_ICCV} use contrastive learning \cite{10209207} to capture robust object features. Since annotations for VIS are costly, \cite{9577716, 9578755, 10.1007/978-3-031-20056-4_19} aim to learn VIS with fewer annotations. 
Offline VIS methods \cite{Athar_Mahadevan20ECCV, 9878426} process videos at the clip level, instead of frame by frame. \cite{guo2024openvis} employs language knowledge to identify novel classes, but lacks the ability to continually learn new knowledge. \cite{9577282} utilizes DETR \cite{10.1007/978-3-030-58452-8_13} to process videos at the clip level, and \cite{cheng2021mask2forme, hwang2021video} use Mask2Former \cite{9878483} to achieve substantial performance gains. 
However, existing VIS methods \cite{Ying_2023_ICCV, Huang_2024_CVPR, 10204941} cannot learn new object categories continually.

\textbf{Continual Learning} (CL) \cite{wei2025compress, 9915459, zhao2021mgsvf, 9156310} enables models to 
consecutively incorporate new information without forgetting. 
Regularization-based CL methods \cite{Kirkpatrick3521, 10555533058903306093, Shmelkov_2017_ICCV} add a penalty term to restrict updates on parameters associated with previous tasks \cite{10555533271443327290, 10.1007/978-3-319-46493-0_37} or distill semantic knowledge from old tasks to new tasks \cite{10.1007/978-3-319-46493-0_37, 10203850, wei2024class, dosovitskiy2021an}. Replay-based CL methods \cite{10555532949963295059, He_2024_CVPR, 10323204_dong} utilize generative networks \cite{NIPS2018_7836, 9878745} to synthesize prior class samples or store some representative images to resolve imbalance between novel and historical categories \cite{9009019, dong2022federated_FCIL}. Architecture-based approaches \cite{yoon2018lifelong, 9156310, Douillard_2022_CVPR} adaptively grow their networks to incorporate new knowledge, while representation-based approaches use prompt tuning \cite{wang2022learning, Smith_2023_CVPR} or contrastive learning \cite{madaan2022representational, 9878593, pham2021dualnet} to explore discriminative features for continual learning. Nevertheless, the aforementioned CL methods \cite{10323204_dong, 9710595, He_2024_CVPR, yoon2018lifelong} are primarily designed for individual images, and cannot explore global video contexts to address video-level catastrophic forgetting.

\begin{figure*}[t]
\centering
\includegraphics[trim = 0mm 40mm 0mm 42mm, clip, width=498pt, height=222pt]
{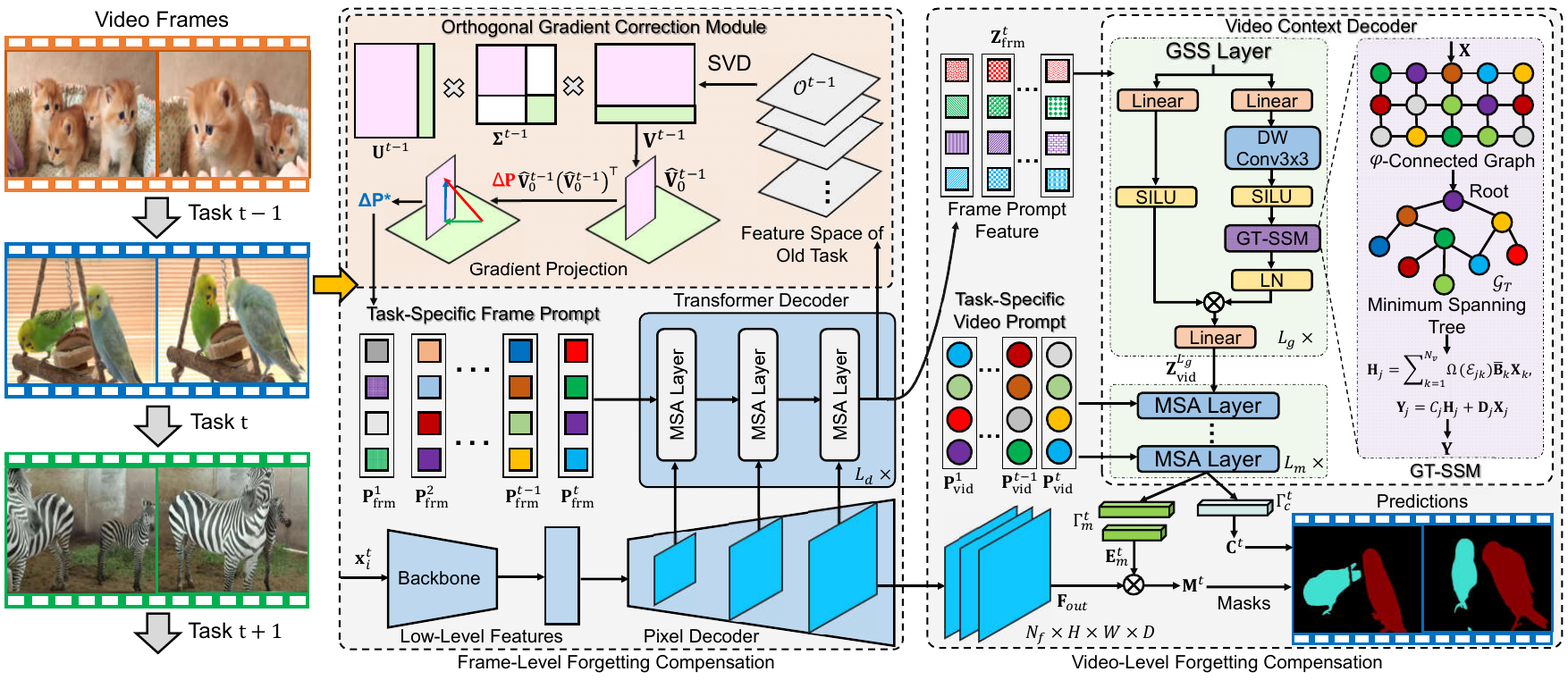}
\vspace{-9mm}
\caption{Algorithmic diagram of the proposed HVPL model. It is composed of a \textit{task-specific frame prompt} and an \textit{orthogonal gradient correction} module to address catastrophic forgetting of old tasks from the frame-level perspective, while also consisting of a \textit{task-specific video prompt} and a \textit{video context decoder} to compensate catastrophic forgetting of old tasks from the video-level perspective. }
\label{fig: overview_pipeline}
\vspace{-5pt}
\end{figure*}

\section{Problem Definition}
In continual video instance segmentation (CVIS), there is a sequence of continuous tasks $\mathcal{T} = \{\mathcal{T}^t\}_{t=1}^T$, where the $t$-th ($t=1, \dots, T$) task $\mathcal{T}^t = \{\mf{x}_i^t, \mf{y}_i^t\}_{i=1}^{N^t}$ consists of $N^t$ video-label pairs. Specifically, the $i$-th video $\mf{x}_i^t\in\mathbb{R}^{N_f\times H\times W\times3}$ contains $N_f$ consecutive frames, and $\mf{y}_i^t \in\mathbb{R}^{N_f\times H\times W\times|\mathcal{Y}^t|}$ represents the corresponding labels. Here, $H$ and $W$ are the height and width of a given video frame. $\mathcal{Y}^t$ denotes the label space of the $t$-th task, and it does not overlap with the label spaces of previous tasks: $\mathcal{Y}^t\cap (\cup_{j=1}^{t-1}\mathcal{Y}^j) = \emptyset$. This indicates that $|\mathcal{Y}^t|$ new classes in the $t$-th task are different from $|\mathcal{Y}^{1:t-1}| = \sum_{j=1}^{t-1}|\mathcal{Y}^j|$ old categories learned from previous tasks. At the $t$-th task, we do not allocate any memory to replay or store training data from old tasks (\emph{i.e.}, rehearsal-free).
Clearly, the lack of training data from previous classes can lead to catastrophic forgetting of old tasks. Upon completing the learning of the $t$-th task in CVIS, we focus on resolving forgetting of old categories to segment and track the $|\mathcal{Y}^{1:t}|$ classes learned so far across frames, including both new and previous classes. To align with real-world CVIS scenarios, we don't require prior knowledge of tasks' data distributions.

\section{The Proposed Model}
The diagram of our model is shown in Fig.~\ref{fig: overview_pipeline}. The proposed HVPL model includes a task-specific frame prompt and an orthogonal gradient correction module to address forgetting at the frame level (Sec.~\ref{sec: frame_level}). Additionally, it incorporates a task-specific video prompt and a video context decoder to alleviate forgetting at the video level (Sec.~\ref{sec: video_level}).

\subsection{Frame-Level Forgetting Compensation}\label{sec: frame_level}
As depicted in Fig.~\ref{fig: overview_pipeline}, we utilize the pretrained Mask2Former \cite{9878483} as the frame-level detector to perform continual video instance segmentation (CVIS) in this paper. It comprises a backbone $\Psi_b$, a pixel decoder $\mathcal{D}_{\mr{pixel}}$ and a Transformer decoder $\mathcal{D}_{\mr{trans}}$, with their network parameters kept frozen to enable continual learning of new tasks. Given a video-label pair $\{\mf{x}_i^t, \mf{y}_i^t\}\in\mathcal{T}^t$ at the $t$-th task, we forward the video $\mf{x}_i^t\in\mathbb{R}^{N_f\times H\times W\times 3}$ into the backbone $\Psi_b$ frame by frame to extract low-level features for each frame. The pixel decoder $\mathcal{D}_{\mr{pixel}}$ then progressively upsamples these low-level features to generate multi-scale per-pixel features. For simplicity, we denote the multi-scale features of the $l$-th ($l=1, \cdots, N_f$) frame in the video $\mf{x}_i^t$ as $\{\mf{F}_{ls}^t \in\mathbb{R}^{H_s\times W_s\times D}\}_{s=1}^S$. Here, $S$ is the number of scales, $H_s$ and $W_s$ are the height and width of the feature at the $s$-th scale. $D$ represents the feature dimension. 
Considering that the queries in Mask2Former \cite{9878483} cannot continuously learn new categories, we propose a learnable task-specific frame prompt $\mf{P}_{\mr{frm}}^t \in\mathbb{R}^{L_p^f\times D}$ to encode the semantic knowledge of new categories in the $t$-th task, where $L_p^f$ denotes the length of frame prompt. For the $l$-th ($l=1, \cdots, N_f$) frame in $\mf{x}_i^t$, we input its multi-scale per-pixel features $\{\mf{F}_{ls}^t\}_{s=1}^S$, along with the frame prompt $\mf{P}_{\mr{frm}}^t$, into the Transformer decoder $\mathcal{D}_{\mr{trans}}$, which includes $L_d$ multi-head self-attention (MSA) layers. The Transformer decoder $\mathcal{D}_{\mr{trans}}$ then outputs the frame prompt features for all frames, denoted as $\mf{Z}_{\mr{frm}}^t = \mathcal{D}_{\mr{trans}}(\mf{x}_i^t, \mf{P}_{\mr{frm}}^t) \in\mathbb{R}^{N_f\times L_p^f\times D}$. It is evident that $\mf{Z}_{\mathrm{frm}}^t$ contains the frame-level global information of object instances in the $t$-th incremental task.

To alleviate forgetting of previously learned tasks, we aim for network updates that preserve the semantic knowledge of old classes under the CVIS setting. In light of this, for any video-label pair $\{\mf{x}_i^{t-1}, \mf{y}_i^{t-1}\} \in\mathcal{T}^{t-1}$ from the $(t{-}1)$-th task, the Transformer decoder $\mathcal{D}_{\mr{trans}}$ should output the same frame prompt features when provided with the task-specific frame prompt $\mf{P}_{\mr{frm}}^{t}$ of the $t$-th task, and the frame prompt $\mf{P}_{\mr{frm}}^{t-1}\!\in\!\mathbb{R}^{L_p^f\times D}$ learned from the $(t{-}1)$-th task as input:
\begin{align}
\mathcal{D}_{\mr{trans}}(\mf{x}_i^{t-1}, \mf{P}_{\mr{frm}}^t) = \mathcal{D}_{\mr{trans}}(\mf{x}_i^{t-1}, \mf{P}_{\mr{frm}}^{t-1}). 
\label{eq: ideal_condition}
\end{align}
Notably, Eq.~\eqref{eq: ideal_condition} is an ideal condition for mitigating catastrophic forgetting of old tasks at the frame level. 
To achieve this, most existing continual learning methods \cite{10203850, Gong_2024_CVPR, chen2024strike} typically perform knowledge distillation between old and new models within individual frames. However, when applied to the CVIS problem, these methods significantly degrade performance on old tasks due to the large appearance variation among different instances of the same category. Furthermore, the multiple task increments further exacerbate the difficulty of knowledge distillation for old classes in CVIS.
To address the above challenges, we develop an orthogonal gradient correction (OGC) module for frame-level forgetting compensation. As presented in Fig.~\ref{fig: overview_pipeline}, this module first constructs orthogonal feature space for the old task and then projects the original gradients used to update the task-specific frame prompt onto the orthogonal feature space of the old task.

\textbf{Construction of Orthogonal Feature Space:}
After utilizing the pixel decoder $\mathcal{D}_{\mr{pixel}}$ to obtain multi-scale per-pixel features $\{\mf{F}_{ls}^{t-1}\}_{s=1}^S$ for the $l$-th frame in the video $\mf{x}_{i}^{t-1}\in\mathcal{T}^{t-1}$, we reshape the feature at the first scale as $\widehat{\mf{F}}_{l1}^{t-1}\in\mathbb{R}^{H_1W_1\times D}$. Considering the parameters of $\mathcal{D}_{\mr{trans}}$ are frozen during training, the proposed OGC module can achieve Eq.~\eqref{eq: ideal_condition} by ensuring that the self-attention between $\widehat{\mf{F}}_{l1}^{t-1}$ and the frame prompt $\mf{P}_{\mr{frm}}^{t}$ is equivalent to the self-attention between $\widehat{\mf{F}}_{l1}^{t-1}$ and $\mf{P}_{\mr{frm}}^{t-1}$. Specifically, given the mapping matrices $\mf{W}_q, \mf{W}_k \in\mathbb{R}^{D\times d}$, the self-attention $\mf{A}_h^t \in\mathbb{R}^{L_p^f\times H_1W_1}$ of the $h$-th head between $\widehat{\mf{F}}_{l1}^{t-1}$ and the frame prompt $\mf{P}_{\mr{frm}}^{t}$ can be formulated as follows:
\begin{align}
\mf{A}_h^t = \sigma(\frac{\mf{P}_{\mr{frm}}^t \mf{W}_q (\widehat{\mf{F}}_{l1}^{t-1}\mf{W}_k)^\top}{\sqrt{d}}), 
\label{eq: self_attention}
\end{align}
where $\sigma$ is the softmax function, and $d$ denotes the dimension of latent feature space. Similarly, we can also express the self-attention of the $h$-th head between $\widehat{\mf{F}}_{l1}^{t-1}$ and $\mf{P}_{\mr{frm}}^{t-1}$ as $\mf{A}_h^{t-1} = \sigma(\frac{\mf{P}_{\mr{frm}}^{t-1} \mf{W}_q (\widehat{\mf{F}}_{l1}^{t-1}\mf{W}_k)^\top}{\sqrt{d}})$. To ensure $\mf{A}_h^t$ and $\mf{A}_h^{t-1}$ are equal, the following condition should be satisfied:
\begin{align}
\mf{P}_{\mr{frm}}^t (\mf{W}_q \mf{W}_k^\top) (\widehat{\mf{F}}_{l1}^{t-1})^\top = \mf{P}_{\mr{frm}}^{t-1} (\mf{W}_q \mf{W}_k^\top) (\widehat{\mf{F}}_{l1}^{t-1})^\top.
\label{eq: attention_condition}
\end{align}

Since $\mf{W}_q$ and $\mf{W}_k$ are frozen weights in the Transformer decoder $\mathcal{D}_{\mr{trans}}$ during training, and $(\mf{W}_q \mf{W}_k^\top)$ is invertible, we can reformulate the condition in Eq.~\eqref{eq: attention_condition} as follows:
\begin{align}
\mf{P}_{\mr{frm}}^t (\widehat{\mf{F}}_{l1}^{t-1})^\top = \mf{P}_{\mr{frm}}^{t-1} (\widehat{\mf{F}}_{l1}^{t-1})^\top.
\label{eq: attention_condition_reformulate}
\end{align}
Then, we expand the task-specific frame prompt $\mf{P}_{\mr{frm}}^t$ of the $t$-th task as $\mf{P}_{\mr{frm}}^t = \mf{P}_{\mr{frm}}^{t-1} + \triangle \mf{P}$, where $\triangle \mf{P} \in\mathbb{R}^{L_p^f\times D}$ is the original gradient used to update the frame prompt when learning the $t$-th task. Thus, 
the condition in Eq.~\eqref{eq: attention_condition_reformulate} can be further expressed as
$\mf{P}_{\mr{frm}}^t (\widehat{\mf{F}}_{l1}^{t-1})^\top = \mf{P}_{\mr{frm}}^{t-1} (\widehat{\mf{F}}_{l1}^{t-1})^\top + \triangle\mf{P}(\widehat{\mf{F}}_{l1}^{t-1})^\top \nonumber = \mf{P}_{\mr{frm}}^{t-1} (\widehat{\mf{F}}_{l1}^{t-1})^\top$.
After eliminating the common term $\mf{P}_{\mr{frm}}^{t-1} (\widehat{\mf{F}}_{l1}^{t-1})^\top$ on both sides, we can obtain:
\begin{align}
\triangle\mf{P}(\widehat{\mf{F}}_{l1}^{t-1})^\top = \mf{0}.
\label{eq: final_condition}
\end{align}

In particular, Eq.~\eqref{eq: final_condition} is the key condition for achieving Eq.~\eqref{eq: ideal_condition}, which can address the catastrophic forgetting of old tasks from the frame-level perspective. However, this condition requires storing the frame-level per-pixel features of all videos from the $(t{-}1)$-th task to alleviate forgetting. It results in high memory consumption, which is impractical in real-world scenarios. To address this limitation, we construct a representative feature space $\mathcal{O}^{t-1} \in\mathbb{R}^{N_o\times D}$ for the $(t{-}1)$-th old task, and replace $\widehat{\mf{F}}_{l1}^{t-1}$ with $\mathcal{O}^{t-1}$ to tackle the forgetting of old tasks at the frame level, where $N_o$ is the number of bases in $\mathcal{O}^{t-1}$. As aforementioned, the outputs of the Transformer decoder $\mathcal{D}_{\mr{trans}}$ hold frame-level global information for each instance in a given video. In light of this, once the learning of the $(t{-}1)$-th task is completed, we randomly sample $B$ videos $\{\mf{x}_i^{t-1}\}_{i=1}^B \in\mathcal{T}^{t-1}$ from the $(t{-}1)$-th task, and utilize the Transformer decoder $\mathcal{D}_{\mr{trans}}$ to extract the frame prompt features $\mf{Z}_{\mr{frm}}^{t-1} = \mathcal{D}_{\mr{trans}}(\mf{x}_i^{t-1}, \mf{P}_{\mr{frm}}^{t-1}) \in\mathbb{R}^{N_f\times L_p^f\times D}$ for each video $\mf{x}_i^{t-1}$. Considering feature diversity, we ensure that the sampled set of $B$ videos, $\{\mf{x}_i^{t-1}\}_{i=1}^B$, fully covers all classes in the $(t{-}1)$-th task. 
To reduce computation and memory costs, we leverage Principal Component Analysis (PCA) \cite{Abdi_cnjdnnPCA} to minimize the feature dimension of $\mf{Z}_{\mr{frm}}^{t-1}$ as $\mathbb{R}^{N_f\times L_p^f\times \frac{D}{N_f}}$. After concatenating all dimensionality-reduced prompt features of $B$ videos as $\mathbb{R}^{B\times N_f\times L_p^f\times \frac{D}{N_f}}$, we reshape this concatenated feature as $\mathcal{O}^{t-1} \in\mathbb{R}^{N_o\times D}$, where $N_o = B\times L_p^f$. Thus, we can substitute $\widehat{\mf{F}}_{l1}^{t-1}$ with $\mathcal{O}^{t-1}$, and transform the condition in Eq.~\eqref{eq: final_condition} as follows:  
\begin{align}
\triangle\mf{P}(\mathcal{O}^{t-1})^\top = \mf{0}.
\label{eq: final_condition_transform}
\end{align}

According to Eq.~\eqref{eq: final_condition_transform}, we conclude that the feature space $\mathcal{O}^{t-1}$ should be constructed during the $(t{-}1)$-th incremental task and stored to support the subsequent learning of the $t$-th task. Since $B\ll N^t$, the memory resource required to store $\mathcal{O}^{t-1}$ consumes less than 0.1M in this paper, which is negligible in real-world applications. 
To achieve Eq.~\eqref{eq: final_condition_transform}, we conduct Singular Value Decomposition (SVD) on $\mathcal{O}^{t-1} = \mf{U}^{t-1} \Sigma^{t-1} (\mf{V}^{t-1})^\top$ to build orthogonal feature space for the $(t{-}1)$-th video instance segmentation task.
Here, $\mf{U}^{t-1}\in\mathbb{R}^{N_o\times N_o}$ and $\mf{V}^{t-1} \in\mathbb{R}^{D\times D}$ are two orthogonal matrices. The main diagonal of matrix $\Sigma^{t-1} \in\mathbb{R}^{N_o\times D}$ contains singular values arranged in descending order. As a result, $\Sigma^{t-1}$ can be actually formulated as follows:  
\begin{align}
\Sigma^{t-1} =
\begin{bmatrix}
\Sigma_1^{t-1} & \mathbf{0} \\
\mathbf{0} & \Sigma_0^{t-1}
\end{bmatrix}
,
\label{eq: definition_sigma}
\end{align}
where $\Sigma_1^{t-1} \in\mathbb{R}^{R\times R}$ denotes the non-zero element of $\Sigma^{t-1}$, while $\Sigma_0^{t-1} \in\mathbb{R}^{(N_o-R)\times (D-R)}$ represents the near-zero element of $\Sigma^{t-1}$. Here, $R$ is the number of non-zero singular values. Similar to Eq.~\eqref{eq: definition_sigma}, we further divide the orthogonal matrix $\mf{V}^{t-1}$ into two parts along its column dimension: $\mf{V}^{t-1} = [\mf{V}_1^{t-1}, \mf{V}_0^{t-1}]$, where the dimensions of $\mf{V}_1^{t-1}$ and $\mf{V}_0^{t-1}$ are $\mathbb{R}^{D\times R}$ and $\mathbb{R}^{D\times (D-R)}$, respectively. Considering the orthogonality of $\mf{V}^{t-1}$ (\emph{i.e.}, $\mf{V}^{t-1}(\mf{V}^{t-1})^\top=\mf{I}$), we transform the SVD on $\mathcal{O}^{t-1} = \mf{U}^{t-1} \Sigma^{t-1} (\mf{V}^{t-1})^\top$ as:
\begin{align}
\mathcal{O}^{t-1}
[\mf{V}_1^{t-1}, \mf{V}_0^{t-1}]
= \mf{U}^{t-1}
\begin{bmatrix}
\Sigma_1^{t-1} & \mathbf{0} \\
\mathbf{0} & \Sigma_0^{t-1}
\end{bmatrix}
.
\label{eq: SVD_transform}
\end{align}
From Eq.~\eqref{eq: SVD_transform}, we can readily observe the following:
\begin{align}
\mathcal{O}^{t-1} \mf{V}_0^{t-1}
= \mf{U}^{t-1}
\begin{bmatrix}
\mathbf{0} \\
\Sigma_0^{t-1}
\end{bmatrix}
\approx \mf{0}.
\label{eq: SVD_transform_observation}
\end{align}

Since $\mf{V}^{t-1}(\mf{V}^{t-1})^\top = \mf{I}$, we define the gradient $\triangle\mf{P}$ used to update the prompt $\mf{P}_{\mr{frm}}^t$ for the $t$-th task as $\triangle\mf{P} = \triangle\mf{P}\mf{V}_0^{t-1}(\mf{V}_0^{t-1})^\top$, leading to the following deduction:
\begin{align}
\triangle\mf{P}(\mathcal{O}^{t-1})^\top = \triangle\mf{P}\mf{V}_0^{t-1}(\mf{V}_0^{t-1})^\top(\mathcal{O}^{t-1})^\top \approx \mf{0}.
\label{eq: final_deduction} 
\end{align}
Particularly, Eq.~\eqref{eq: final_deduction} enables us to effectively satisfy the condition in Eq.~\eqref{eq: final_condition_transform} when we consider $\mf{V}_0^{t{-}1}$ as the orthogonal feature space of the $(t{-}1)$-th task for gradient projection.

\textbf{Gradient Projection:}
If we strictly satisfy the requirement in Eq.~\eqref{eq: final_deduction} to update the frame prompt $\mf{P}_{\mr{frm}}^t$ for the $t$-th task, our model can effectively address the forgetting of old classes. However, this strict requirement may hinder the learning on the $t$-th task to some extent. To balance learning the $t$-th task and alleviating forgetting of old tasks, we propose an elastic threshold $\xi\in[0,1]$ to determine the orthogonal feature space of the $(t{-}1)$-th task. Specifically, we use $\xi$ to split $\mf{V}^{t-1}\in\mathbb{R}^{D\times D}$ into two parts: $\widehat{\mf{V}}_1^{t-1}\in\mathbb{R}^{D\times \xi D}$ and $\widehat{\mf{V}}_0^{t-1}\in\mathbb{R}^{D\times (1-\xi) D}$. Then we regard $\widehat{\mf{V}}_0^{t-1}$ as the orthogonal feature space of the $(t{-}1)$-th task and use it to replace $\mf{V}_0^{t-1}$ in Eq.~\eqref{eq: final_deduction} for gradient projection. During training, we first project the original gradient $\triangle\mf{P}$ on $\widehat{\mf{V}}_0^{t-1}$ to obtain the projected gradient $\triangle\mf{P}^*$. Then we utilize $\triangle\mf{P}^*$ to update the frame prompt $\mf{P}_{\mr{frm}}^t$, effectively addressing forgetting at the frame level. As a result, $\triangle\mf{P}^*$ can be formulated as:
\begin{align}
\triangle\mf{P}^* = \triangle\mf{P}\widehat{\mf{V}}_0^{t-1}(\widehat{\mf{V}}_0^{t-1})^\top.
\label{eq: final_gradient} 
\end{align}

\subsection{Video-Level Forgetting Compensation}\label{sec: video_level}
Although the proposed task-specific frame prompt and orthogonal gradient correction module in Sec.~\ref{sec: frame_level} alleviate forgetting at the frame level, they overlook intrinsic relationships between different instances across frames. Moreover, they lack an understanding of the global semantic context of the entire video, which makes it challenging to address the CVIS problem. To address these challenges, as shown in Fig.~\ref{fig: overview_pipeline}, we propose a learnable task-specific video prompt $\mf{P}_{\mr{vid}}^t\in\mathbb{R}^{L_p^v\times D}$ and a video context decoder $\mathcal{D}_{\mr{video}}$, alleviating forgetting of old tasks at the video level, where $L_p^v$ denotes the length of video prompt. The video context decoder $\mathcal{D}_{\mr{video}}$ consists of $L_g$ graph-guided state space (GSS) layers to capture structural category information across frames, and $L_m$ multi-head self-attention (MSA) layers \cite{dosovitskiy2021an} to explore task-specific global video contexts for the $t$-th task.

\textbf{GSS Layer:} 
After the Transformer decoder $\mathcal{D}_{\mr{trans}}$ outputs the frame prompt features $\mf{Z}_{\mr{frm}}^t = \mathcal{D}_{\mr{trans}}(\mf{x}_i^t, \mf{P}_{\mr{frm}}^t) \in\mathbb{R}^{N_f\times L_p^f\times D}$ for any video $\mf{x}_i^t \in\mathcal{T}^t$ at the $t$-th task, we reshape $\mf{Z}_{\mr{frm}}^t$ as $\mf{Z}_{\mr{vid}}^0 \in\mathbb{R}^{N_v\times D}$, and feed $\mf{Z}_{\mr{vid}}^0$ into the GSS layers in $\mathcal{D}_{\mr{video}}$, where $N_v = N_f\times L_p^f$. In the $l$-th ($l = 1, \cdots, L_g$) GSS layer, the feature $\mf{Z}_{\mr{vid}}^{l-1}$ outputted by the $(l{-}1)$-th GSS layer is forwarded into two branches. As shown in Fig.~\ref{fig: overview_pipeline}, the right branch utilizes a linear projection, a depth-wise convolution, and a SILU function to encode $\mf{Z}_{\mr{vid}}^{l-1}$ as $\mf{X}\in\mathbb{R}^{N_v\times D}$. Then $\mf{X}$ is fed into the graph-traversal state space module $\mathrm{GT}\text{-}\mathrm{SSM}(\cdot)$ to capture structural category information across frames. The final feature $\mf{Z}_{\mr{vid}}^{l} \in\mathbb{R}^{N_v\times D}$, extracted by the $l$-th GSS layer, is formally defined as:
\begin{align}
\mf{Z}_{\mr{vid}}^{l} = \mathbf{W}_p\big(\mathrm{LN}(\mathrm{GT}\text{-}\mathrm{SSM}(\mathbf{X})) \otimes \mathbf{Z}_{\mathrm{left}} \big),
\label{eq: feature_GSS}
\end{align}
where $\mathbf{Z}_{\mathrm{left}} \in\mathbb{R}^{N_v\times D}$ is the feature encoded by the left branch, including a linear projection and a SILU function. $\otimes$ is the element-wise multiplication operation. $\mathbf{W}_p\in\mathbb{R}^{D\times D}$ and $\mathrm{LN}$ denote the linear mapping and layer normalization.

\begin{algorithm}[t]
\small
\caption{Determination of $\{\mf{H}_j\}_{j=1}^{N_v}$. }
\LinesNumbered
\label{alg: determination_H_i}
\textbf{Initialize:} Use \cite{kahn1962topological} to obtain the breadth-first topological order (BTO) of $\mathcal{G}_T$: $(\mr{Root}, \cdots, \mr{Leaf})\leftarrow \mr{BTO}(\mathcal{G}_T)$;

\For{$j~\mr{in}~(\mr{Leaf}, \cdots, \mr{Root})$}
{
$\zeta_j = \overline{\mf{B}}_j\mf{X}_j + \sum_{s\in\{r|\mr{Par}(r)=j\}} \overline{\mf{A}}_s\zeta_s$;
}

\For{$j~\mr{in}~(\mr{Root}, \cdots, \mr{Leaf})$}
{

\If{$j~\mr{is}~\mr{Root}$}
{
$\mf{H}_j = \zeta_j$; \\
}
\Else{
$\mf{H}_j = \overline{\mf{A}}_j (\mf{H}_{\mr{Par}(j)} - \overline{\mf{A}}_j \zeta_j) + \zeta_j$. \\
}

}

\end{algorithm}

\begin{table*}[t]
\centering
\setlength{\tabcolsep}{1.1mm}
\caption{Comparison experiments on OVIS \cite{qi2021occluded} under the 15-5 and 15-10 settings. }
\vspace{-4mm}
\resizebox{1.0\linewidth}{!}{
\begin{tabular}{l|ccccccc|ccccccc}
\toprule
\makecell[c]{\multirow{2}{*}{Comparison Methods}} & \multicolumn{7}{c|}{15-5 (3 Tasks)} & \multicolumn{7}{c}{15-10 (2 Tasks)} \\ 
& \#Params & $\mr{AP}$ & $\mr{AP}_{50}$ & $\mr{AP}_{75}$ & $\mr{AR}_1$& $\mr{FAP}$& $\mr{FAR}_{1}$ & \#Params & $\mr{AP}$ & $\mr{AP}_{50}$ & $\mr{AP}_{75}$ & $\mr{AR}_1$ & $\mr{FAP}$ & $\mr{FAR}_{1}$ \\
\hline
Finetuning & 53.37M &\ \ 3.47  &\ \ 8.22 &\ \ 2.34 & \ \ 3.06 & 38.41 & 33.08 & 53.12M &\ \ 5.12 & 11.93 & \ \ 3.78 &\ \ 3.76 & 75.90 & 79.81 \\
MiB \cite{cermelli2020modeling} (CVPR'2020) & 41.92M &\ \ 3.88 &\ \ 8.94 &\ \ 2.91 &\ \ 3.39 & 36.62 &36.19 & 41.92M &\ \ 5.07 & 11.70 &\ \ 3.51 &\ \ 3.85 & 70.29 & 74.43 \\
CoMFormer \cite{10203850} (CVPR'2023) & 41.92M &\ \ 3.72 &\ \ 9.29 &\ \ 2.61 &\ \ 3.10 & 33.31 & 34.31 & 41.92M &\ \ 5.76 & 13.68 &\ \ 4.06 &\ \ 3.88 & 54.21 & 63.66 \\ 

PLOP \cite{douillard2021plop} + NeST \cite{10.1007/978-3-031-73347-5_11} ~~[ECCV'2024] & 41.94M &\ \ 4.07 &\ \ 9.54 &\ \ 3.35 &\ \ 3.16 & 34.07 & 34.14 & 41.96M &\ \ 4.79 & 11.88 &\ \ 3.10 &\ \ 3.81 & 66.70 & 75.85 \\ 

CoMFormer \cite{10203850} + NeST \cite{10.1007/978-3-031-73347-5_11} (ECCV'2024) & 41.94M &\ \ 4.44 & 10.90 &\ \ 3.14 &\ \ 3.37 & 32.25 & 35.47 & 41.96M &\ \ 5.87 & 14.58 &\ \ 4.05 &\ \ 4.44 & 51.20 & 58.36  \\

BalConpas \cite{chen2024strike} (ECCV'2024) & 41.92M &\ \ 2.42 &\ \ 6.17 &\ \ 1.48 &\ \ 2.37 & 51.25 &58.63 & 41.92M &\ \ 3.88 &\ \ 9.47 &\ \ 2.52 &\ \ 3.36 & 83.05 & 82.88  \\

ECLIPSE \cite{Kim_2024_CVPR} (CVPR'2024) & \ \ 0.47M &\ \ 6.62 &15.05  &\ \ 4.60 &\ \ 5.29 &28.08 &34.29 &\ \ 0.47M &\ \ 6.82 &15.63  &\ \ 4.69 &\ \ 4.97 &22.11 &25.01 \\
\hline
\rowcolor{lightgray}
\textbf{HVPL} (\textbf{Ours}) & \ \ 0.92M & \textbf{\textcolor{deepred}{11.09}} & \textbf{\textcolor{deepred}{23.57}} & \ \ \textbf{\textcolor{deepred}{9.87}} & \ \ \textbf{\textcolor{deepred}{8.77}} & \ \ \textbf{\textcolor{deepred}{9.56}} & \textbf{\textcolor{deepred}{10.74}} & \  \ 0.92M  & \textbf{\textcolor{deepred}{11.92}} & \textbf{\textcolor{deepred}{25.85}} & \textbf{\textcolor{deepred}{10.60}} & \ \ \textbf{\textcolor{deepred}{8.69}} & \textbf{\textcolor{deepred}{16.64}} & \textbf{\textcolor{deepred}{14.86}} \\

\bottomrule
\end{tabular}}
\label{tab: compa_ovis}
\vspace{-3mm}
\end{table*}

\textbf{GT-SSM:} Inspired by vision Mamba \cite{liu2024vmamba}, we define a set of learnable variables $\{\mf{A}_j\in\mathbb{R}^{\mathcal{Q}\times \mathcal{Q}}, \mf{B}_j\in\mathbb{R}^{\mathcal{Q}\times D}, \mf{C}_j\in\mathbb{R}^{D\times \mathcal{Q}}, \mf{D}_j\in\mathbb{R}\}_{j=1}^{N_v}$, and a timescale parameter $\{\triangle_j\}_{j=1}^{N_v}$. Then we employ the zero-order hold rule \cite{gu2023mamba} to obtain the discrete variables $\{\overline{\mf{A}}_j\in\mathbb{R}^{\mathcal{Q}\times \mathcal{Q}}, \overline{\mf{B}}_j\in\mathbb{R}^{\mathcal{Q}\times D}\}_{j=1}^{N_v}$:
\begin{align}
\!\! \overline{\mf{A}}_j = \exp(\triangle_j\mf{A}_j), \overline{\mf{B}}_j = (\exp(\triangle_j\mf{A}_j) - I) \mf{A}_j^{-1}\mf{B}_j, \!
\label{eq: mamba_discrete_variables}
\end{align}
where $\mathcal{Q}$ is the dimension of hidden states. 
Different from traditional vision Mamba \cite{liu2024vmamba, vimicml2024} with fixed traversal path, we propose a graph-traversal state space module $\mathrm{GT}\text{-}\mathrm{SSM}(\cdot)$ to explore inherent structural inter-class relationships across frames. As presented in Fig.~\ref{fig: overview_pipeline}, we use the input feature $\mf{X} \in\mathbb{R}^{N_v\times D}$ to construct an undirected $\varphi$-connected graph $\mathcal{G} = (\mathcal{V}, \mathcal{E})$, where each vertex has $\varphi$ neighbors, $\mathcal{V}$ contains $N_v$ vertices, and $\mathcal{E}$ denotes the edges between them, which are computed using the cosine similarity between adjacent vertices. To eliminate edges with significant dissimilarity, we apply the contractive boruvka algorithm \cite{georgiadis2016contractive} on $\mathcal{G}$ to obtain a minimum spanning tree $\mathcal{G}_T$. Subsequently, we traverse all vertices in $\mathcal{G}_T$ to obtain the latent feature $\mf{H}_j\in\mathbb{R}^{\mathcal{Q}\times 1}$ for the $j$-the ($j=1, \cdots, N_v$) hidden state: 
\begin{align}
\!\mf{H}_j = \sum\nolimits_{k=1}^{N_v} \Omega(\mathcal{E}_{jk}) \overline{B}_k\mf{X}_k,~
\Omega(\mathcal{E}_{jk}) = \Pi_{m\in\Phi_{jk}} \overline{A}_m,\! 
\label{eq: hidden_state_tree}
\end{align}
where $\mf{X}_k\in\mathbb{R}^{D\times 1}$ is the $k$-th row of $\mf{X}$, and $\Phi_{jk}$ denotes the index set of the hyperedge $\mathcal{E}_{jk}$ that traces from the $j$-th vertex to the $k$-th vertex. 
Since the complexity of using Eq.~\eqref{eq: hidden_state_tree} to obtain $\{\mf{H}_j\}_{j=1}^{N_v}$ is $\mathcal{O}(N_v^2)$, we design a dynamic traversal strategy (Algorithm~\ref{alg: determination_H_i}), reducing its complexity to $\mathcal{O}(N_v)$. In Algorithm~\ref{alg: determination_H_i}, $\mr{Par}(\cdot)$ is used to the find parent node. After computing $\{\mf{H}_j\}_{j=1}^{N_v}$ via Algorithm~\ref{alg: determination_H_i}, we can obtain $\mf{Y}_j = \mf{C}_j\mf{H}_j+\mf{D}_j\mf{X}_j \in\mathbb{R}^{D\times 1}$. Subsequently, we concatenate $\{\mf{Y}_j\}_{j=1}^{N_v}$ as $\mf{Y} = \mathrm{GT}\text{-}\mathrm{SSM}(\mathbf{X}) \in\mathbb{R}^{N_v\times D}$, which is regarded as the output of the proposed GT-SSM.

\textbf{MSA Layer:}
As aforementioned, the $L_g$-th GSS layer can output the feature $\mf{Z}_{\mr{vid}}^{L_g}\in\mathbb{R}^{N_v\times D}$, which is embedded with structural inter-class relations across frames. After denoting $\mf{Z}_{\mr{vid}}^{L_g}$ as $\mf{F}_{\mr{vid}}^0$, we 
forward it along with the task-specific video prompt $\mf{P}_{\mr{vid}}^t$ into $L_m$ MSA layers to capture global video contexts of the $t$-th task from $\mf{F}_{\mr{vid}}^0$, and propagate relevant semantic information to the video prompt $\mf{P}_{\mr{vid}}^t$. For the $l$-th ($l=1, \cdots, L_m$) MSA layer in the video context decoder $\mathcal{D}_{\mr{video}}$, we formulate the output $\Phi_h \in\mathbb{R}^{L_p^v\times d}$ of the self-attention at the $h$-th head as follows: 
\begin{align}
\Phi_h = \sigma(\frac{\mf{P}_{\mr{vid}}^t \mf{W}_q (\mf{F}_{\mr{vid}}^{l-1} \mf{W}_k)^\top}{\sqrt{d}})(\mf{F}_{\mr{vid}}^{l-1} \mf{W}_v), 
\label{eq: self_attention_video_prompt}
\end{align}
where $\mf{W}_q, \mf{W}_k, \mf{W}_v \in\mathbb{R}^{D\times d}$ denote the projection matrices. $\mf{F}_{\mr{vid}}^{l-1} \in\mathbb{R}^{N_v\times D}$ denotes the latent feature extracted by the $(l{-}1)$-th MSA layer. 
Subsequently, we concatenate the outputs of $H$ heads along channel dimension and encode this concatenated feature via a projection matrix $\mf{W}_o \in\mathbb{R}^{D\times D}$ ($D=dH$) to achieve $\mr{MSA}(\mf{P}_{\mr{vid}}^t, \mf{F}_{\mr{vid}}^{l-1}) \!=\! \mf{P}_{\mr{vid}}^t \!+\! \mr{Concat}(\Phi_1, \cdots\!,\Phi_H)\mf{W}_o$.
The video prompt feature $\mf{F}_{\mr{vid}}^l \in\mathbb{R}^{L_p^v\times D}$ outputted by the $l$-th MSA layer is: 
\begin{align}
\!\!\!\! \mf{F}_{\mr{vid}}^l \!=\! \mr{MSA}(\mf{P}_{\mr{vid}}^t, \mf{F}_{\mr{vid}}^{l-1}) \!+\! \mr{MLP}(\mr{MSA}(\mf{P}_{\mr{vid}}^t, \mf{F}_{\mr{vid}}^{l-1})),\!\!
\label{eq: MSHA_video_prompt_first_layer}
\end{align} 
where MLP is the multi-layer perceptron. Then, the $L_m$-th MSA layer outputs the final video prompt feature $\mf{F}_{\mr{vid}}^{L_m} \in\mathbb{R}^{L_p^v\times D}$, which has captured global video contexts of the $t$-th task from $\mf{Z}_{\mr{frm}}^t$ to alleviate forgetting at the video level.

\begin{table*}[t]
\centering
\setlength{\tabcolsep}{1.1mm}
\caption{Comparison experiments on YouTube-VIS 2021 \cite{9008283} under the 30-10 and 20-4 settings. }
\vspace{-4mm}
\resizebox{1.0\linewidth}{!}{
\begin{tabular}{l|ccccccc|ccccccc}
\toprule
\makecell[c]{\multirow{2}{*}{Comparison Methods}} & \multicolumn{7}{c|}{30-10 (2 Tasks)} & \multicolumn{7}{c}{20-4 (6 Tasks)} \\ 
& \#Params & $\mr{AP}$ & $\mr{AP}_{50}$ & $\mr{AP}_{75}$ & $\mr{AR}_1$& $\mr{FAP}$& $\mr{FAR}_{1}$ & \#Params & $\mr{AP}$ & $\mr{AP}_{50}$ & $\mr{AP}_{75}$ & $\mr{AR}_1$ & $\mr{FAP}$ & $\mr{FAR}_{1}$ \\
\hline
Finetuning & 53.13M & 21.50 & 33.63 & 23.57 &24.54 & 81.19 & 81.10 & 54.13M & 11.03 & 16.07 & 12.38 & 12.75 & 29.65 & 29.98 \\
MiB \cite{cermelli2020modeling} (CVPR'2020) & 41.92M & 24.66 & 38.25 & 27.49 & 27.79 & 65.55 & 63.72 & 41.92M & 13.43 & 19.99 & 15.22 & 14.95 & 30.81 & 29.67 \\
CoMFormer \cite{10203850} (CVPR'2023) & 41.92M & 25.96 & 40.58 & 28.53 & 29.41 & 58.44 & 55.36 & 41.92M & 15.25 & 22.96 & 17.09 & 16.76 & 29.94 & 30.12 \\ 

PLOP \cite{douillard2021plop} + NeST \cite{10.1007/978-3-031-73347-5_11} ~~[ECCV'2024] & 42.00M & 25.79 & 40.52 & 28.48 & 28.99 & 59.46 & 59.06 & 41.96M & 12.73 & 19.30 & 16.12 & 13.90 & 27.97 & 27.53 \\ 

CoMFormer \cite{10203850} + NeST \cite{10.1007/978-3-031-73347-5_11} (ECCV'2024) & 42.00M & 28.70 & 43.99 & 31.78 & 31.81 & 46.27 & 45.46 & 41.96M & 15.37 & 22.93 & 17.45 & 16.67 & 30.94 & 30.71  \\

BalConpas \cite{chen2024strike} (ECCV'2024) & 41.92M & 22.12 & 34.31 & 24.35 & 25.39 & 76.72 &75.03 & 41.92M & 10.87 & 15.55 & 12.90 & 11.87 &32.84 & 32.67  \\

ECLIPSE \cite{Kim_2024_CVPR} (CVPR'2024) & \ \ 0.47M & 29.68 & 43.86 & 33.28 & 32.29  &40.59 & 37.75 & \ \ 0.46M & 30.52 & 43.06 & 34.49 & 32.02 & 22.83 & 22.47 \\
\hline
\rowcolor{lightgray}
\textbf{HVPL} (\textbf{Ours}) & \ \ 0.92M & \textbf{\textcolor{deepred}{43.24}} & \textbf{\textcolor{deepred}{60.13}} & \textbf{\textcolor{deepred}{47.50}} & \textbf{\textcolor{deepred}{40.32}} & \textbf{\textcolor{deepred}{15.69}} & \textbf{\textcolor{deepred}{14.83}} & \  \ 0.92M & \textbf{\textcolor{deepred}{35.05}} & \textbf{\textcolor{deepred}{45.79}} & \textbf{\textcolor{deepred}{39.17}} & \textbf{\textcolor{deepred}{34.19}} & \textbf{\textcolor{deepred}{20.42}} & \textbf{\textcolor{deepred}{20.15}} \\

\bottomrule
\end{tabular}}
\label{tab: compa_YouTube_2021}
\vspace{-4mm}
\end{table*}

\begin{table*}[t]
\centering
\setlength{\tabcolsep}{1.1mm}
\caption{Comparison experiments on YouTube-VIS 2019 \cite{9008283} under the 20-2 and 20-5 settings. }
\vspace{-4mm}
\resizebox{1.0\linewidth}{!}{
\begin{tabular}{l|ccccccc|ccccccc}
\toprule
\makecell[c]{\multirow{2}{*}{Comparison Methods}} & \multicolumn{7}{c|}{20-2 (11 Tasks)} & \multicolumn{7}{c}{20-5 (5 Tasks)} \\ 
& \#Params & $\mr{AP}$ & $\mr{AP}_{50}$ & $\mr{AP}_{75}$ & $\mr{AR}_1$& $\mr{FAP}$& $\mr{FAR}_{1}$ & \#Params & $\mr{AP}$ & $\mr{AP}_{50}$ & $\mr{AP}_{75}$ & $\mr{AR}_1$ & $\mr{FAP}$ & $\mr{FAR}_{1}$ \\
\hline
Finetuning & 55.32M &\ \ 7.35 &11.31 & \ \ 7.93 &\ \ 9.31 & 19.66 & 19.52 & 53.82M & 15.34 & 23.71 & 16.34 & 18.18 & 35.81 & 34.99 \\
MiB \cite{cermelli2020modeling} (CVPR'2020) & 41.92M &\ \ 9.05 & 14.24 &\ \ 9.57 & 11.44 & 19.63 & 19.57 & 41.92M & 18.16 & 28.42 & 19.34 & 20.71 & 34.68 & 34.78 \\
CoMFormer \cite{10203850} (CVPR'2023) & 41.92M  &\ \ 9.58 & 14.80 & 10.21 & 12.96 &17.14 &17.16 & 41.92M & 20.93 & 33.24 & 22.17 & 25.43 & 30.62 & 30.42  \\ 

PLOP \cite{douillard2021plop} + NeST \cite{10.1007/978-3-031-73347-5_11} ~~[ECCV'2024] & 41.94M &\ \ 9.45 & 14.94 &\ \ 9.94 & 12.25 & 19.32 & 19.96 & 41.97M & 18.83 & 29.62 & 19.89 & 22.05 &33.28 & 32.61  \\ 

CoMFormer \cite{10203850} + NeST \cite{10.1007/978-3-031-73347-5_11} (ECCV'2024)& 41.94M & 10.29 & 16.33 & 10.81 & 14.31 & 17.32 & 17.24 & 41.97M & 21.39 & 33.65 & 23.01 & 25.29 & 33.64 & 33.14  \\

BalConpas \cite{chen2024strike} (ECCV'2024) & 41.92M  &\ \ 6.42 & 10.06 &\ \ 6.80 &\ \ 7.80 & 20.06 &20.03
& 41.92M & 14.04 & 21.71 & 14.93 & 16.31 &39.51 & 39.32  \\

ECLIPSE \cite{Kim_2024_CVPR} (CVPR'2024) & \ \ 0.45M & 30.24 &45.27 & 33.43 & 34.00 & 13.86 & 13.63 & \ \ 0.47M &32.62 &48.35 &35.99 &38.10 &20.29 &19.22  \\
\hline
\rowcolor{lightgray}
\textbf{HVPL} (\textbf{Ours}) & \ \ 0.90M & \textbf{\textcolor{deepred}{31.68}} & \textbf{\textcolor{deepred}{45.75}} & \textbf{\textcolor{deepred}{34.11}} & \textbf{\textcolor{deepred}{34.79}} & \textbf{\textcolor{deepred}{10.67}} & \ \ \textbf{\textcolor{deepred}{9.93}} & \ \ 0.92M & \textbf{\textcolor{deepred}{34.79}} & \textbf{\textcolor{deepred}{49.55}} & \textbf{\textcolor{deepred}{37.59}} & \textbf{\textcolor{deepred}{38.53}} & \textbf{\textcolor{deepred}{19.12}} & \textbf{\textcolor{deepred}{17.26}} \\

\bottomrule
\end{tabular}}
\label{tab: compa_YouTube_2019}
\vspace{-1mm}
\end{table*}

Following \cite{hwang2021video, VITA}, we employ a classifier head and a mask head on the top of the video prompt feature $\mf{F}_{\mr{vid}}^{L_m}$ for predictions. As presented in Fig.~\ref{fig: overview_pipeline}, the classifier head $\Gamma_c^t\in\mathbb{R}^{D\times (|\mathcal{Y}^t|+1)}$ uses a single linear layer to predict the class probability $\mf{C}^t = \sigma(\mf{F}_{\mr{vid}}^{L_m} \Gamma_c^t) \in\mathbb{R}^{L_p^v\times (|\mathcal{Y}^t|+1)}$, where $\sigma$ is the softmax function.  $(|\mathcal{Y}^t|+1)$ is the number of new classes along with an auxiliary category, ``no object''. The mask head $\Gamma_m^t\in\mathbb{R}^{D\times D}$ is a two-layer MLP, which converts $\mf{F}_{\mr{vid}}^{L_m}$ to mask embeddings $\mf{E}_m^t = \mf{F}_{\mr{vid}}^{L_m}\Gamma_m^t\in\mathbb{R}^{L_p^v\times D}$. As introduced in Sec.~\ref{sec: frame_level}, the pixel decoder $\mathcal{D}_{\mr{pixel}}$ can output multi-scale features $\{\mf{F}_{ls}^t \in\mathbb{R}^{H_s\times W_s\times D} \}_{s=1}^S$ for the $l$-th ($l=1, \cdots, N_f$) frame in the video $\mf{x}_i^t$. Then we upsample the features at the final scale for all frames to $\mathbb{R}^{H\times W\times D}$ and concatenate them along the temporal axis, resulting in $\mf{F}_{\mr{out}}\in\mathbb{R}^{N_f\times H\times W\times D}$. The predicted mask $\mf{M}^t \in\mathbb{R}^{N_f\times L_p^v\times H\times W}$ for each video prompt can be obtained by the matrix multiplication between $\mf{F}_{\mr{out}}$ and $\mf{E}_m^t$.

\textbf{Training and Testing:} For training, we use the objective loss proposed in \cite{VITA} to optimize our HVPL model. 
When the learning of the $t$-th task is finished, we can utilize all frame and video prompts $\{\mf{P}_{\mr{frm}}^j, \mf{P}_{\mr{vid}}^j\}_{j=1}^t$ learned so far for testing. Firstly, we concatenate all frame prompts as $\mf{P}_{\mr{frm}}^* = [\mf{P}_{\mr{frm}}^1; \cdots; \mf{P}_{\mr{frm}}^t] \in\mathbb{R}^{tL_p^f\times D}$, and all video prompts as $\mf{P}_{\mr{vid}}^* = [\mf{P}_{\mr{vid}}^1; \cdots; \mf{P}_{\mr{vid}}^t] \in\mathbb{R}^{tL_p^v\times D}$. Secondly, a given test video $\mf{x}$ and $\mf{P}_{\mr{frm}}^*$ are forwarded into the frame-level detector to obtain the frame prompt feature $\mf{Z}_{\mr{frm}}^*$, which is then fed into $L_g$ GSS layers in $\mathcal{D}_{\mr{video}}$ to produce the feature $\mf{Z}_{\mr{vid}}^*$. Finally, we input $\mf{Z}_{\mr{vid}}^*$ and $\mf{P}_{\mr{vid}}^*$ into $L_m$ MSA layers in $\mathcal{D}_{\mr{video}}$ to output the video prompt feature $\mf{F}_{\mr{vid}}^*$, and utilize the classifier and mask heads for predictions.

\section{Experiments}

\subsection{Experimental Settings}
\textbf{Datasets:}
We verify the effectiveness of our model on three video instance segmentation datasets (\emph{i.e.}, YouTube-VIS 2019 \cite{9008283}, YouTube-VIS 2021 \cite{9008283}, and OVIS \cite{qi2021occluded}) under the CVIS setting.  
On \textbf{YouTube-VIS 2019} \cite{9008283}, we consider 20-2 and 20-5 settings. The 20-2 and 20-5 settings indicate first learning 20 classes, followed by ten continual tasks, each with 2 new classes ($T=11$); and followed by 4 tasks, each with 5 new classes ($T=5$). 
On \textbf{YouTube-VIS 2021} \cite{9008283}, we set 30-10 and 20-4. The settings of 30-10 and 20-4 involve learning 30 classes followed by one task with 10 new classes ($T=2$), and learning 20 classes followed by 5 continual tasks with 4 new classes each ($T=6$). 
On \textbf{OVIS} \cite{qi2021occluded}, we set 15-5 and 15-10 for the CVIS problem. Here, 15-5 and 15-10 involve learning 15 classes followed by 2 task with 5 new classes each ($T=3$), and learning 15 classes followed by one task with 10 new classes ($T=2$).

\textbf{Implementation Details:}
For the network architecture, we introduce Mask2Former \cite{9878483} as the frame-level detector, where the backbone is ResNet-50 \cite{7780459}. The video context decoder $\mathcal{D}_{\mr{video}}$ consists of 6 GSS layers and 3 MSA layers.
For the lengths of task-specific frame and video prompts, we set $L_p^f=100, L_p^v=100$ for the first VIS task and set $L_p^f=10, L_p^v=10$ to learn the $t$-th ($t\geq2$) VIS task. We adopt the same preprocessing strategy for input video frames during training and testing as proposed in \cite{VITA}.
Besides, we set $\xi=0.7$ to determine the orthogonal feature space $\widehat{\mf{V}}_0^{t-1}$ for the $(t-1)$-th task. 
For fair comparisons, all comparative methods employ the same network architecture and data augmentation. We utilize the Adam optimizer to train our model, where the initial learning rate is $5.0\times 10^{-5}$. 

\begin{table}[t]
\centering
\setlength{\tabcolsep}{1.1mm}
\caption{Comparisons of each incremental task ($t\geq 2$) on YouTube-VIS 2021 \cite{9008283} under the 20-4 setting. }
\vspace{-3mm}
\resizebox{1.0\linewidth}{!}{
\begin{tabular}{l|ccccc}
\toprule
\makecell[c]{Comparison Methods}
& $t=2$ & $t=3$ & $t=4$ & $t=5$ & $t=6$  \\
\hline
Finetuning &\ \ 3.94 & \ \ 4.22 & \ \ 3.68 &\ \ 3.39 &\ \ 1.51  \\
MiB \cite{cermelli2020modeling} (CVPR'2020) &\ \ 9.51 	&\ \ 7.79 	&\ \ 6.77 	&\ \ 4.75 	&\ \ 2.34 
 \\
CoMFormer \cite{10203850} (CVPR'2023) & 11.85 	&12.28 	&\ \ 8.70 	&\ \ 5.85 	&\ \ 3.39 
    \\ 

PLOP \cite{douillard2021plop} + NeST \cite{10.1007/978-3-031-73347-5_11} ~~[ECCV'2024] &\ \ 9.65 	&\ \ 7.22 	&\ \ 4.70 	&\ \ 3.68 	&\ \ 1.69 
  \\ 

CoMFormer \cite{10203850} + NeST \cite{10.1007/978-3-031-73347-5_11} (ECCV'2024) & 14.68 	&11.82 	&\ \ 7.73 	&\ \ 5.69 	&\ \ 2.86 
 \\

BalConpas \cite{chen2024strike} (ECCV'2024) &\ \ 5.30 	&\ \ 3.84 	&\ \ 3.37 	&\ \ 1.90 	&\ \ 1.39 
  \\

ECLIPSE \cite{Kim_2024_CVPR} (CVPR'2024) & 33.69 	&27.56 	&24.19 	&23.86 	&22.76 
 \\
\hline
\rowcolor{lightgray}
\textbf{HVPL} (\textbf{Ours}) & \textbf{\textcolor{deepred}{41.83}} & \textbf{\textcolor{deepred}{34.06}} & \textbf{\textcolor{deepred}{29.09}} & \textbf{\textcolor{deepred}{25.62}}  & \textbf{\textcolor{deepred}{22.81}}   \\

\bottomrule
\end{tabular}}
\label{tab: compa_YouTube_2021_20_4_task_wise}
\vspace{-2mm}
\end{table}

\textbf{Evaluation Metrics:}
Following \cite{VITA, hwang2021video, Ying_2023_ICCV}, we utilize the average precision (AP) and average recall (AR) at different Intersection over Union (IoU) thresholds (\emph{i.e.}, $\mr{AP}, \mr{AP}_{50}, \mr{AR}_{1}$) to assess the results of video instance segmentation. Moreover, we leverage the forgetting rates of $\mr{AP}$ and $\mr{AR}_1$ (\emph{i.e.}, FAP and $\mr{FAR}_1$) to assess the performance of our model in tackling forgetting. Here, we define FAP as: 
$\mr{FAP}= \frac{1}{|\mathcal{Y}^{1:T{-}1}|} \sum_{t=1}^{T-1} (\frac{1}{T-t} \sum_{k=1}^{|\mathcal{Y}^t|}(\mathcal{A}_k^t-\mathcal{A}_k^T)/\mathcal{A}_k^t)$. $\mathcal{A}_k^t$ and $\mathcal{A}_k^T$ denote the AP tested for the $k$-th class when it is first learned and the AP tested after completing the final task, respectively. Notably, $\mathrm{FAR}_1$ is formulated similarly to FAP.

\begin{figure*}[t]
\centering
\includegraphics[trim = 0mm 54mm 0mm 54mm, clip, width=498pt, height=172pt]
{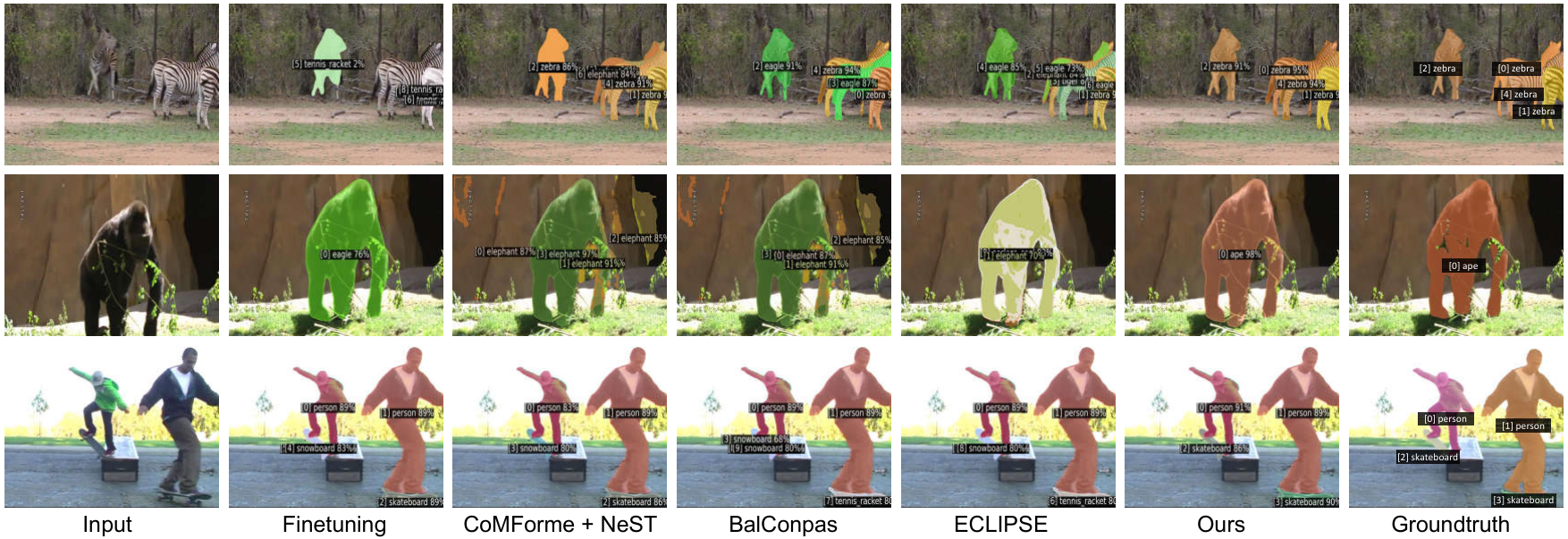}
\vspace{-21pt}
\caption{Comparison results of some selected video frames from YouTube-VIS 2019 \cite{9008283} under the 20-5 setting (zoom in for a better view). }
\label{fig: vis_YouTube2019_30-5}
\vspace{-3mm}
\end{figure*}

\subsection{Performance Comparison}
As shown in Tabs.~\ref{tab: compa_ovis}--\ref{tab: compa_YouTube_2021_20_4_task_wise}, we introduce extensive comparisons across diverse CVIS settings to illustrate the efficacy of our model. The proposed model significantly surpasses the prompt learning-based method (\emph{i.e.}, ECLIPSE \cite{Kim_2024_CVPR}) in all evaluation metrics. This large improvement demonstrates the superiority of our model in handling the CVIS problem by addressing forgetting at both the frame and video levels. The performance of our HVPL is superior than the knowledge distillation-based methods (\emph{e.g.}, CoMFormer \cite{10203850}, NeST \cite{10.1007/978-3-031-73347-5_11} and BalConpas \cite{chen2024strike}). It verifies that the task-specific video prompt and video context decoder can effectively capture global video contexts of old tasks to alleviate forgetting. Besides, our model has a comparable number of trainable parameters (\#Params) to that of ECLIPSE \cite{Kim_2024_CVPR}, verifying its efficiency in addressing the CVIS problem. 
Fig.~\ref{fig: vis_YouTube2019_30-5} visualizes some selected video frames from YouTube-VIS 2019 \cite{9008283} to further exhibit the competitive performance of our HVPL.

\begin{table}[t]
\centering
\footnotesize
\setlength{\tabcolsep}{1.76mm}
\renewcommand{\arraystretch}{1.04}
\caption{Ablation studies on YouTube-VIS 2021 \cite{9008283}.  }
\vspace{-3mm}
\resizebox{1.0\linewidth}{!}{
\begin{tabular}{c|l|cccc|c}
\toprule
\multicolumn{2}{c|}{Variants} & Base & \makecell[c]{Base+TFP} & \makecell[c]{Base+TFP\\ ~~~\quad+TVP} & \makecell[c]{Base+TFP+\\ ~TVP+GSS}& ~\textbf{Ours}~ \\
\hline 
\multirow{4}{*}{\rotatebox{90}{Modules}}
& TFP & \xmarkg & \cmark & \cmark & \cmark & \cmark \\
& TVP  & \xmarkg & \xmarkg & \cmark & \cmark & \cmark \\
& GSS & \xmarkg & \xmarkg & \xmarkg & \cmark & \cmark \\
& OGC & \xmarkg & \xmarkg & \xmarkg & \xmarkg & \cmark \\
\hline
\multirow{6}{*}{\rotatebox{90}{30-10}}
& $\mr{AP}$ &27.53 & 30.02 & 39.07 &40.37 & \textcolor{deepred}{\textbf{43.24}} \\
& $\mr{AP}_{50}$ &40.24 &42.82 & 54.11 &55.62 & \textcolor{deepred}{\textbf{60.13}} \\
& $\mr{AP}_{75}$ &29.70 &32.85 & 43.83 & 44.62& \textcolor{deepred}{\textbf{47.50}} \\
& $\mr{AR}_{1}$ & 26.87 &29.26 & 35.66 & 37.50 & \textcolor{deepred}{\textbf{40.32}} \\
& $\mr{FAP}$ & 96.48 & 74.88 & 31.05 & 30.40 &  \textcolor{deepred}{\textbf{15.69}} \\
& $\mr{FAR}_{1}$ & 95.79 & 70.88 & 29.51 &28.93 &  \textcolor{deepred}{\textbf{14.83}} \\

\bottomrule
\end{tabular}
}
\label{tab: ablation_YouTube_2021}
\vspace{-4mm}
\end{table}

\subsection{Ablation Study}
To demonstrate the efficacy of each module, as shown in Tabs.~\ref{tab: ablation_YouTube_2021}, we conduct ablation studies about the task-specific frame prompt (TFP), GSS layers in $\mathcal{D}_{\mr{video}}$, task-specific video prompt (TVP) and orthogonal gradient correction (OGC) module. ``Base'' indicates the performance of our model without the TFP, GSS, TVP and OGC modules, achieved by continually finetuning Mask2Former with new tasks. Notably, since the MSA layers in the video context decoder $\mathcal{D}_{\mr{video}}$ and TVP must be bound together to alleviate video-level forgetting, ablation studies on TVP generally include the MSA layers as well. 
The ablation experiments reveal a significant improvement in our model's performance when all modules collaboratively handle the CVIS problem. It demonstrates that the task-specific frame prompt and orthogonal gradient correction effectively alleviate forgetting at the frame level. The task-specific video prompt alleviates forgetting from the video-level perspective by incorporating video context decoder to capture global video contexts.

\subsection{Analysis of Orthogonal Feature Space}
\begin{figure}[t]
\centering
\includegraphics[trim = 6mm 45mm 6mm 45mm, clip, width=236pt, height=96pt]
{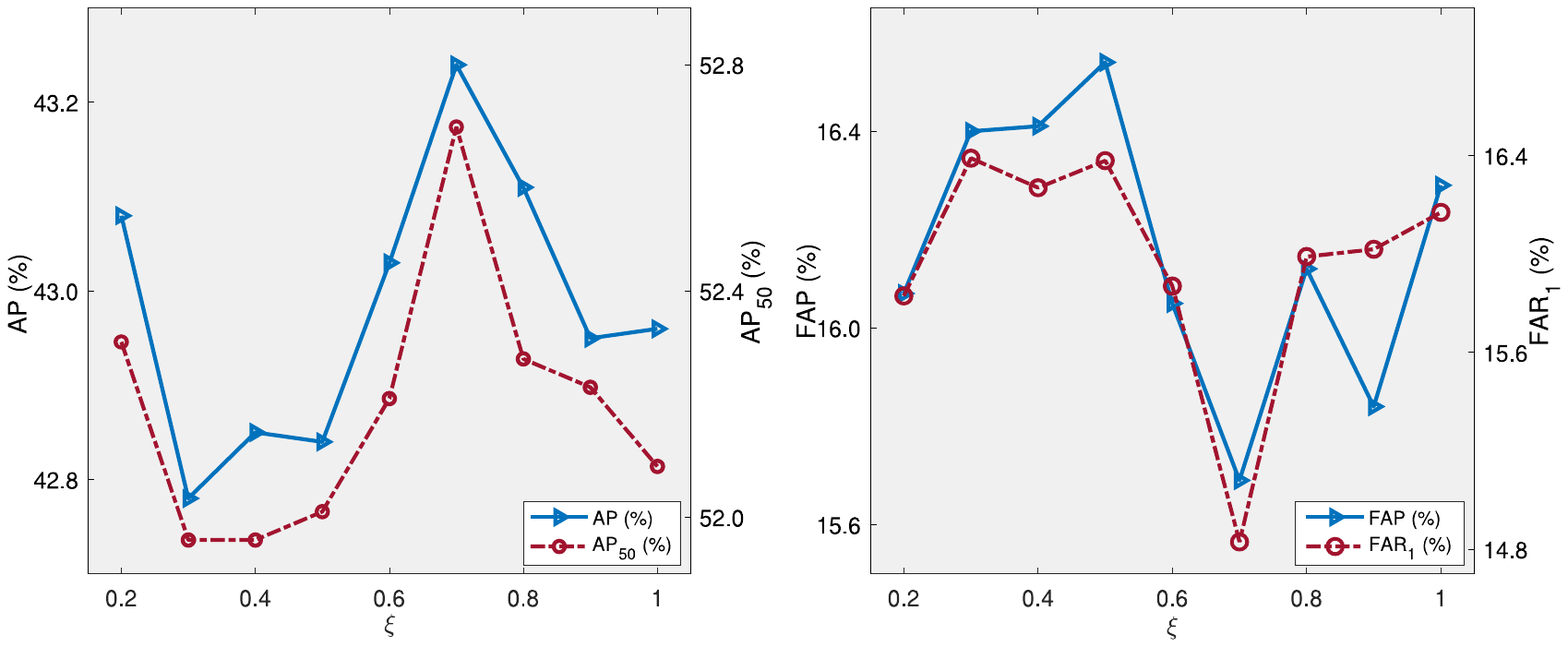}
\vspace{-9mm}
\caption{Analysis of the orthogonal feature space on YouTube-VIS 2021 \cite{9008283} under the 30-10 setting. }
\label{fig: analysis_orthogonal_space}
\vspace{-4mm}
\end{figure}

This subsection investigates the effect of orthogonal feature space $\widehat{\mf{V}}_0^{t-1}$ on the performance of our HVPL model by applying different thresholds $\xi\in[0,1]$. As introduced in Fig.~\ref{fig: analysis_orthogonal_space}, our model presents stable performance in addressing the CVIS problem across a range of $\xi$ values from 0.2 to 1.0. Besides, the proposed model has the optimal results on YouTube-VIS 2021 \cite{9008283} under the 30-10 setting when we use $\xi=0.7$ to determine the orthogonal feature space $\widehat{\mf{V}}_0^{t-1}$ for the old task. As a result, we set $\xi=0.7$ for all comparison experiments in this paper. Additionally, these parameter experiments validate the effectiveness of $\xi$ in balancing the forgetting of old tasks and the learning of new tasks.

\vspace{-1mm}
\section{Conclusion}
\vspace{-1mm}
In this work, we introduce a new problem referred to as Continual Video Instance Segmentation (CVIS), and propose a Hierarchical Visual Prompt Learning (HVPL) model to resolve it by mitigating catastrophic forgetting from both frame-level and video-level perspectives. 
Specifically, we design a task-specific frame prompt and an orthogonal gradient correction module to alleviate forgetting at the frame level by projecting the frame prompt gradients onto the orthogonal feature space of old classes. Furthermore, we develop a task-specific video prompt and a video context decoder to alleviate forgetting at the video level by exploring task-specific global video contexts. Experiments exhibit the efficacy of our HVPL model in handling the CVIS problem. 
In the future, we will incorporate large language models \cite{li2025system, zhang2024mm} with our HVPL to tackle large-scale continual tasks.

\newpage
{
    \small
    \bibliographystyle{ieeenat_fullname}
    \bibliography{main}

\begin{thebibliography}{87}
\providecommand{\natexlab}[1]{#1}
\providecommand{\url}[1]{\texttt{#1}}
\expandafter\ifx\csname urlstyle\endcsname\relax
  \providecommand{\doi}[1]{doi: #1}\else
  \providecommand{\doi}{doi: \begingroup \urlstyle{rm}\Url}\fi

\bibitem[Abati et~al.(2020)Abati, Tomczak, Blankevoort, Calderara, Cucchiara, and Bejnordi]{9156310}
Davide Abati, Jakub Tomczak, Tijmen Blankevoort, Simone Calderara, Rita Cucchiara, and Babak~Ehteshami Bejnordi.
\newblock Conditional channel gated networks for task-aware continual learning.
\newblock In \emph{CVPR}, pages 3930--3939, 2020.

\bibitem[Abdi and Williams(2010)]{Abdi_cnjdnnPCA}
Hervé Abdi and Lynne~J. Williams.
\newblock Principal component analysis.
\newblock \emph{WIREs Computational Statistics}, 2\penalty0 (4):\penalty0 433--459, 2010.

\bibitem[Athar et~al.(2020)Athar, Mahadevan, O{\v{s}}ep, Leal-Taix{\'e}, and Leibe]{Athar_Mahadevan20ECCV}
Ali Athar, Sabarinath Mahadevan, Aljo{\v{s}}a O{\v{s}}ep, Laura Leal-Taix{\'e}, and Bastian Leibe.
\newblock Stem-seg: Spatio-temporal embeddings for instance segmentation in videos.
\newblock In \emph{ECCV}, 2020.

\bibitem[Belouadah and Popescu(2019)]{9009019}
Eden Belouadah and Adrian Popescu.
\newblock Il2m: Class incremental learning with dual memory.
\newblock In \emph{ICCV}, pages 583--592, 2019.

\bibitem[Carion et~al.(2020)Carion, Massa, Synnaeve, Usunier, Kirillov, and Zagoruyko]{10.1007/978-3-030-58452-8_13}
Nicolas Carion, Francisco Massa, Gabriel Synnaeve, Nicolas Usunier, Alexander Kirillov, and Sergey Zagoruyko.
\newblock End-to-end object detection with transformers.
\newblock In \emph{ECCV}, page 213–229, 2020.

\bibitem[Cermelli et~al.(2020)Cermelli, Mancini, Rota~Bul\`o, Ricci, and Caputo]{cermelli2020modeling}
Fabio Cermelli, Massimiliano Mancini, Samuel Rota~Bul\`o, Elisa Ricci, and Barbara Caputo.
\newblock Modeling the background for incremental learning in semantic segmentation.
\newblock In \emph{CVPR}, 2020.

\bibitem[Cermelli et~al.(2023)Cermelli, Cord, and Douillard]{10203850}
Fabio Cermelli, Matthieu Cord, and Arthur Douillard.
\newblock Comformer: Continual learning in semantic and panoptic segmentation.
\newblock In \emph{CVPR}, pages 3010--3020, 2023.

\bibitem[Chen et~al.(2024)Chen, Cong, Luo, Ip, and Kwong]{chen2024strike}
Jinpeng Chen, Runmin Cong, Yuxuan Luo, Horace Ho~Shing Ip, and Sam Kwong.
\newblock Strike a balance in continual panoptic segmentation.
\newblock In \emph{ECCV}, 2024.

\bibitem[Cheng et~al.(2021)Cheng, Choudhuri, Misra, Kirillov, Girdhar, and Schwing]{cheng2021mask2forme}
Bowen Cheng, Anwesa Choudhuri, Ishan Misra, Alexander Kirillov, Rohit Girdhar, and Alexander~G. Schwing.
\newblock Mask2former for video instance segmentation.
\newblock \emph{arxiv preprint arxiv: 2112.10764}, 2021.

\bibitem[Cheng et~al.(2022)Cheng, Misra, Schwing, Kirillov, and Girdhar]{9878483}
Bowen Cheng, Ishan Misra, Alexander~G. Schwing, Alexander Kirillov, and Rohit Girdhar.
\newblock Masked-attention mask transformer for universal image segmentation.
\newblock In \emph{CVPR}, pages 1280--1289, 2022.

\bibitem[Cheng et~al.(2024)Cheng, Oh, Price, Lee, and Schwing]{cheng2023putting}
Ho~Kei Cheng, Seoung~Wug Oh, Brian Price, Joon-Young Lee, and Alexander Schwing.
\newblock Putting the object back into video object segmentation.
\newblock In \emph{CVPR}, 2024.

\bibitem[Choudhuri et~al.(2023)Choudhuri, Chowdhary, and Schwing]{10204984}
Anwesa Choudhuri, Girish Chowdhary, and Alexander~G. Schwing.
\newblock Context-aware relative object queries to unify video instance and panoptic segmentation.
\newblock In \emph{CVPR}, pages 6377--6386, 2023.

\bibitem[Ding et~al.(2023)Ding, Liu, He, Jiang, Torr, and Bai]{MOSE_ding}
Henghui Ding, Chang Liu, Shuting He, Xudong Jiang, Philip~HS Torr, and Song Bai.
\newblock {MOSE}: A new dataset for video object segmentation in complex scenes.
\newblock In \emph{ICCV}, 2023.

\bibitem[Dong et~al.(2022)Dong, Wang, Fang, Sun, Xu, Wang, and Zhu]{dong2022federated_FCIL}
Jiahua Dong, Lixu Wang, Zhen Fang, Gan Sun, Shichao Xu, Xiao Wang, and Qi Zhu.
\newblock Federated class-incremental learning.
\newblock In \emph{CVPR}, 2022.

\bibitem[Dong et~al.(2024)Dong, Li, Cong, Sun, Zhang, and Van~Gool]{10323204_dong}
Jiahua Dong, Hongliu Li, Yang Cong, Gan Sun, Yulun Zhang, and Luc Van~Gool.
\newblock No one left behind: Real-world federated class-incremental learning.
\newblock \emph{IEEE Transactions on Pattern Analysis and Machine Intelligence}, 46\penalty0 (4):\penalty0 2054--2070, 2024.

\bibitem[Dosovitskiy et~al.(2021)Dosovitskiy, Beyer, Kolesnikov, Weissenborn, Zhai, Unterthiner, Dehghani, Minderer, Heigold, Gelly, Uszkoreit, and Houlsby]{dosovitskiy2021an}
Alexey Dosovitskiy, Lucas Beyer, Alexander Kolesnikov, Dirk Weissenborn, Xiaohua Zhai, Thomas Unterthiner, Mostafa Dehghani, Matthias Minderer, Georg Heigold, Sylvain Gelly, Jakob Uszkoreit, and Neil Houlsby.
\newblock An image is worth 16x16 words: Transformers for image recognition at scale.
\newblock In \emph{ICLR}, 2021.

\bibitem[Douillard et~al.(2021)Douillard, Chen, Dapogny, and Cord]{douillard2021plop}
Arthur Douillard, Yifu Chen, Arnaud Dapogny, and Matthieu Cord.
\newblock Plop: Learning without forgetting for continual semantic segmentation.
\newblock In \emph{CVPR}, 2021.

\bibitem[Douillard et~al.(2022)Douillard, Ram\'e, Couairon, and Cord]{Douillard_2022_CVPR}
Arthur Douillard, Alexandre Ram\'e, Guillaume Couairon, and Matthieu Cord.
\newblock Dytox: Transformers for continual learning with dynamic token expansion.
\newblock In \emph{CVPR}, pages 9285--9295, 2022.

\bibitem[Fan et~al.(2022)Fan, Zhang, and Tan]{10.1007/978-3-031-20056-4_19}
Junsong Fan, Zhaoxiang Zhang, and Tieniu Tan.
\newblock Pointly-supervised panoptic segmentation.
\newblock In \emph{ECCV}, pages 319--336, Cham, 2022. Springer Nature Switzerland.

\bibitem[Fini et~al.(2022)Fini, Da~Costa, Alameda-Pineda, Ricci, Alahari, and Mairal]{9878593}
Enrico Fini, Victor G.~Turrisi Da~Costa, Xavier Alameda-Pineda, Elisa Ricci, Karteek Alahari, and Julien Mairal.
\newblock Self-supervised models are continual learners.
\newblock In \emph{CVPR}, pages 9611--9620, 2022.

\bibitem[Fischer et~al.(2023)Fischer, Huang, Pang, Qiu, Chen, Darrell, and Yu]{10209207}
Tobias Fischer, Thomas~E. Huang, Jiangmiao Pang, Linlu Qiu, Haofeng Chen, Trevor Darrell, and Fisher Yu.
\newblock Qdtrack: Quasi-dense similarity learning for appearance-only multiple object tracking.
\newblock \emph{IEEE Transactions on Pattern Analysis and Machine Intelligence}, 45\penalty0 (12):\penalty0 15380--15393, 2023.

\bibitem[Fu et~al.(2021)Fu, Liu, Iqbal, De~Mello, Shi, and Kautz]{9577716}
Yang Fu, Sifei Liu, Umar Iqbal, Shalini De~Mello, Humphrey Shi, and Jan Kautz.
\newblock Learning to track instances without video annotations.
\newblock In \emph{CVPR}, pages 8676--8685, 2021.

\bibitem[Georgiadis and Koutsoukos(2016)]{georgiadis2016contractive}
Loukas Georgiadis and Xenofon Koutsoukos.
\newblock The contractive boruvka algorithm for minimum spanning trees.
\newblock \emph{Journal of Algorithms}, 94:\penalty0 1--17, 2016.

\bibitem[Ghiasi et~al.(2021)Ghiasi, Cui, Srinivas, Qian, Lin, Cubuk, Le, and Zoph]{9578639}
Golnaz Ghiasi, Yin Cui, Aravind Srinivas, Rui Qian, Tsung-Yi Lin, Ekin~D. Cubuk, Quoc~V. Le, and Barret Zoph.
\newblock Simple copy-paste is a strong data augmentation method for instance segmentation.
\newblock In \emph{CVPR}, pages 2917--2927, 2021.

\bibitem[Gong et~al.(2024)Gong, Yu, Wang, and Xiao]{Gong_2024_CVPR}
Yizheng Gong, Siyue Yu, Xiaoyang Wang, and Jimin Xiao.
\newblock Continual segmentation with disentangled objectness learning and class recognition.
\newblock In \emph{CVPR}, pages 3848--3857, 2024.

\bibitem[Griffin et~al.(2020)Griffin, Florence, and Corso]{9093335}
Brent~A. Griffin, Victoria Florence, and Jason~J. Corso.
\newblock Video object segmentation-based visual servo control and object depth estimation on a mobile robot.
\newblock In \emph{WACV}, pages 1636--1646, 2020.

\bibitem[Gu and Dao(2023)]{gu2023mamba}
Albert Gu and Tri Dao.
\newblock Mamba: Linear-time sequence modeling with selective state spaces.
\newblock \emph{arXiv preprint arXiv:2312.00752}, 2023.

\bibitem[Guo et~al.(2024)Guo, Huang, He, Liu, Xiao, Chen, and Zhang]{guo2024openvis}
Pinxue Guo, Tony Huang, Peiyang He, Xuefeng Liu, Tianjun Xiao, Zhaoyu Chen, and Wenqiang Zhang.
\newblock Openvis: Open-vocabulary video instance segmentation.
\newblock \emph{International Journal of Computer Vision}, 2024.

\bibitem[Han et~al.(2022)Han, Hwang, Oh, Park, Kim, Kim, and Kim]{9879517}
Su~Ho Han, Sukjun Hwang, Seoung~Wug Oh, Yeonchool Park, Hyunwoo Kim, Min-Jung Kim, and Seon~Joo Kim.
\newblock Visolo: Grid-based space-time aggregation for efficient online video instance segmentation.
\newblock In \emph{CVPR}, pages 2886--2895, 2022.

\bibitem[Hao et~al.(2024)Hao, Peng, Yawei, Xinxin, and Xiankai]{fang2024unified}
Fang Hao, Wu Peng, Li Yawei, Zhang Xinxin, and Lu Xiankai.
\newblock Unified embedding alignment for open-vocabulary video instance segmentation.
\newblock In \emph{ECCV}, 2024.

\bibitem[He(2024)]{He_2024_CVPR}
Jiangpeng He.
\newblock Gradient reweighting: Towards imbalanced class-incremental learning.
\newblock In \emph{CVPR}, pages 16668--16677, 2024.

\bibitem[He et~al.(2016)He, Zhang, Ren, and Sun]{7780459}
Kaiming He, Xiangyu Zhang, Shaoqing Ren, and Jian Sun.
\newblock Deep residual learning for image recognition.
\newblock In \emph{CVPR}, pages 770--778, 2016.

\bibitem[Heo et~al.(2022)Heo, Hwang, Oh, Lee, and Kim]{VITA}
Miran Heo, Sukjun Hwang, Seoung~Wug Oh, Joon-Young Lee, and Seon~Joo Kim.
\newblock Vita: Video instance segmentation via object token association.
\newblock In \emph{NeurIPS}, 2022.

\bibitem[Heo et~al.(2023)Heo, Hwang, Hyun, Kim, Oh, Lee, and Kim]{10204941}
Miran Heo, Sukjun Hwang, Jeongseok Hyun, Hanjung Kim, Seoung~Wug Oh, Joon-Young Lee, and Seon~Joo Kim.
\newblock A generalized framework for video instance segmentation.
\newblock In \emph{CVPR}, pages 14623--14632, 2023.

\bibitem[Huang et~al.(2022)Huang, Yu, and Anandkumar]{huang2022minvis}
De-An Huang, Zhiding Yu, and Anima Anandkumar.
\newblock Minvis: A minimal video instance segmentation framework without video-based training.
\newblock In \emph{NeurIPS}, 2022.

\bibitem[Huang et~al.(2024)Huang, Huang, Yu, Lan, Radhakrishnan, Alvarez, Shrivastava, and Anandkumar]{Huang_2024_CVPR}
Shuaiyi Huang, De-An Huang, Zhiding Yu, Shiyi Lan, Subhashree Radhakrishnan, Jose~M. Alvarez, Abhinav Shrivastava, and Anima Anandkumar.
\newblock What is point supervision worth in video instance segmentation?
\newblock In \emph{CVPR Workshops}, pages 2671--2681, 2024.

\bibitem[Hwang et~al.(2021)Hwang, Heo, Oh, and Kim]{hwang2021video}
Sukjun Hwang, Miran Heo, Seoung~Wug Oh, and Seon~Joo Kim.
\newblock Video instance segmentation using inter-frame communication transformers.
\newblock \emph{NeurIPS}, 34:\penalty0 13352--13363, 2021.

\bibitem[Kahn(1962)]{kahn1962topological}
Arthur~B. Kahn.
\newblock Topological sorting of large networks.
\newblock \emph{Communications of the ACM}, 5\penalty0 (11):\penalty0 558--562, 1962.

\bibitem[Ke et~al.(2021)Ke, Li, Danelljan, Tai, Tang, and Yu]{ke2021prototypical}
Lei Ke, Xia Li, Martin Danelljan, Yu-Wing Tai, Chi-Keung Tang, and Fisher Yu.
\newblock Prototypical cross-attention networks for multiple object tracking and segmentation.
\newblock In \emph{NeurIPS}, 2021.

\bibitem[Ke et~al.(2023)Ke, Danelljan, Ding, Tai, Tang, and Yu]{maskfreevis}
Lei Ke, Martin Danelljan, Henghui Ding, Yu-Wing Tai, Chi-Keung Tang, and Fisher Yu.
\newblock Mask-free video instance segmentation.
\newblock In \emph{CVPR}, 2023.

\bibitem[Kim et~al.(2024)Kim, Yu, and Hwang]{Kim_2024_CVPR}
Beomyoung Kim, Joonsang Yu, and Sung~Ju Hwang.
\newblock Eclipse: Efficient continual learning in panoptic segmentation with visual prompt tuning.
\newblock In \emph{CVPR}, pages 3346--3356, 2024.

\bibitem[Kim et~al.(2025)Kim, Kang, Heo, Hwang, Oh, and Kim]{10.1007/978-3-031-72667-5_6}
Hanjung Kim, Jaehyun Kang, Miran Heo, Sukjun Hwang, Seoung~Wug Oh, and Seon~Joo Kim.
\newblock Visage: Video instance segmentation with appearance-guided enhancement.
\newblock In \emph{ECCV}, pages 93--109, 2025.

\bibitem[Kirkpatrick et~al.(2017)Kirkpatrick, Pascanu, Rabinowitz, Veness, Desjardins, Rusu, Milan, Quan, Ramalho, Grabska-Barwinska, Hassabis, Clopath, Kumaran, and Hadsell]{Kirkpatrick3521}
James Kirkpatrick, Razvan Pascanu, Neil Rabinowitz, Joel Veness, Guillaume Desjardins, Andrei~A. Rusu, Kieran Milan, John Quan, Tiago Ramalho, Agnieszka Grabska-Barwinska, Demis Hassabis, Claudia Clopath, Dharshan Kumaran, and Raia Hadsell.
\newblock Overcoming catastrophic forgetting in neural networks.
\newblock \emph{Proceedings of the National Academy of Sciences}, 114\penalty0 (13):\penalty0 3521--3526, 2017.

\bibitem[Li et~al.(2023)Li, Yu, Rao, Zhou, and Lu]{Li_2023_ICCV}
Junlong Li, Bingyao Yu, Yongming Rao, Jie Zhou, and Jiwen Lu.
\newblock Tcovis: Temporally consistent online video instance segmentation.
\newblock In \emph{ICCV}, pages 1097--1107, 2023.

\bibitem[Li et~al.(2024)Li, Ding, Yuan, Zhang, Pang, Cheng, Chen, Liu, and Loy]{SegSurvey}
Xiangtai Li, Henghui Ding, Haobo Yuan, Wenwei Zhang, Jiangmiao Pang, Guangliang Cheng, Kai Chen, Ziwei Liu, and Chen~Change Loy.
\newblock Transformer-based visual segmentation: A survey.
\newblock \emph{TPAMI}, 2024.

\bibitem[Li and Hoiem(2016)]{10.1007/978-3-319-46493-0_37}
Zhizhong Li and Derek Hoiem.
\newblock Learning without forgetting.
\newblock In \emph{ECCV}, pages 614--629, 2016.

\bibitem[Li et~al.(2025)Li, Zhang, Zhang, Zhang, Liu, et~al.]{li2025system}
Zhong-Zhi Li, Duzhen Zhang, Ming-Liang Zhang, Jiaxin Zhang, Zengyan Liu, et~al.
\newblock {From system 1 to system 2: A survey of reasoning large language models}.
\newblock \emph{arXiv preprint arXiv:2502.17419}, 2025.

\bibitem[Liu et~al.(2021)Liu, Ramanathan, Mahajan, Yuille, and Yang]{9578755}
Qing Liu, Vignesh Ramanathan, Dhruv Mahajan, Alan Yuille, and Zhenheng Yang.
\newblock Weakly supervised instance segmentation for videos with temporal mask consistency.
\newblock In \emph{CVPR}, pages 13963--13973, 2021.

\bibitem[Liu et~al.(2024)Liu, Tian, Zhao, Yu, Xie, Wang, Ye, Jiao, and Liu]{liu2024vmamba}
Yue Liu, Yunjie Tian, Yuzhong Zhao, Hongtian Yu, Lingxi Xie, Yaowei Wang, Qixiang Ye, Jianbin Jiao, and Yunfan Liu.
\newblock {VM}amba: Visual state space model.
\newblock In \emph{NeurIPS}, 2024.

\bibitem[Madaan et~al.(2022)Madaan, Yoon, Li, Liu, and Hwang]{madaan2022representational}
Divyam Madaan, Jaehong Yoon, Yuanchun Li, Yunxin Liu, and Sung~Ju Hwang.
\newblock Representational continuity for unsupervised continual learning.
\newblock In \emph{ICLR}, 2022.

\bibitem[Masana et~al.(2022)Masana, Liu, Twardowski, Menta, Bagdanov, and van~de Weijer]{9915459}
Marc Masana, Xialei Liu, Bartłomiej Twardowski, Mikel Menta, Andrew~D. Bagdanov, and Joost van~de Weijer.
\newblock Class-incremental learning: Survey and performance evaluation on image classification.
\newblock \emph{IEEE Transactions on Pattern Analysis and Machine Intelligence}, 1\penalty0 (1):\penalty0 1--20, 2022.

\bibitem[Mohamed et~al.(2021)Mohamed, Ewaisha, Siam, Rashed, Yogamani, Hamdy, El-Dakdouky, and El-Sallab]{9575445}
Eslam Mohamed, Mahmoud Ewaisha, Mennatullah Siam, Hazem Rashed, Senthil Yogamani, Waleed Hamdy, Mohamed El-Dakdouky, and Ahmad El-Sallab.
\newblock Monocular instance motion segmentation for autonomous driving: Kitti instancemotseg dataset and multi-task baseline.
\newblock In \emph{2021 IEEE Intelligent Vehicles Symposium (IV)}, pages 114--121, 2021.

\bibitem[Onsori-Wechtitsch and Bailer(2022)]{10024449}
Stefanie Onsori-Wechtitsch and Werner Bailer.
\newblock Multi-head instance segmentation of indoor scenes for ar/dr applications.
\newblock In \emph{AIVR}, pages 114--118, 2022.

\bibitem[Pham et~al.(2021)Pham, Liu, and HOI]{pham2021dualnet}
Quang Pham, Chenghao Liu, and Steven HOI.
\newblock Dualnet: Continual learning, fast and slow.
\newblock In \emph{NeurIPS}, 2021.

\bibitem[Qi et~al.(2021)Qi, Gao, Hu, Wang, Liu, Bai, Belongie, Yuille, Torr, and Bai]{qi2021occluded}
Jiyang Qi, Yan Gao, Yao Hu, Xinggang Wang, Xiaoyu Liu, Xiang Bai, Serge Belongie, Alan Yuille, Philip Torr, and Song Bai.
\newblock Occluded video instance segmentation: Dataset and {ICCV} 2021 challenge.
\newblock In \emph{NeurIPS Datasets and Benchmarks Track}, 2021.

\bibitem[Rebuffi et~al.(2017)Rebuffi, Kolesnikov, Sperl, and Lampert]{Rebuffi_2017_CVPR}
Sylvestre-Alvise Rebuffi, Alexander Kolesnikov, Georg Sperl, and Christoph~H. Lampert.
\newblock icarl: Incremental classifier and representation learning.
\newblock In \emph{CVPR}, 2017.

\bibitem[Ren et~al.(2015)Ren, He, Girshick, and Sun]{NIPS2015_14bfa6bb}
Shaoqing Ren, Kaiming He, Ross Girshick, and Jian Sun.
\newblock Faster r-cnn: Towards real-time object detection with region proposal networks.
\newblock In \emph{NeurIPS}, 2015.

\bibitem[Ritter et~al.(2018)Ritter, Botev, and Barber]{10555533271443327290}
Hippolyt Ritter, Aleksandar Botev, and David Barber.
\newblock Online structured laplace approximations for overcoming catastrophic forgetting.
\newblock In \emph{NeurIPS}, page 3742–3752. Curran Associates Inc., 2018.

\bibitem[Shin et~al.(2017)Shin, Lee, Kim, and Kim]{10555532949963295059}
Hanul Shin, Jung~Kwon Lee, Jaehong Kim, and Jiwon Kim.
\newblock Continual learning with deep generative replay.
\newblock In \emph{NeurIPS}, page 2994–3003. Curran Associates Inc., 2017.

\bibitem[Shmelkov et~al.(2017)Shmelkov, Schmid, and Alahari]{Shmelkov_2017_ICCV}
Konstantin Shmelkov, Cordelia Schmid, and Karteek Alahari.
\newblock Incremental learning of object detectors without catastrophic forgetting.
\newblock In \emph{ICCV}, 2017.

\bibitem[Smith et~al.(2023)Smith, Karlinsky, Gutta, Cascante-Bonilla, Kim, Arbelle, Panda, Feris, and Kira]{Smith_2023_CVPR}
James~Seale Smith, Leonid Karlinsky, Vyshnavi Gutta, Paola Cascante-Bonilla, Donghyun Kim, Assaf Arbelle, Rameswar Panda, Rogerio Feris, and Zsolt Kira.
\newblock Coda-prompt: Continual decomposed attention-based prompting for rehearsal-free continual learning.
\newblock In \emph{CVPR}, pages 11909--11919, 2023.

\bibitem[Strudel et~al.(2021)Strudel, Garcia, Laptev, and Schmid]{9710959}
Robin Strudel, Ricardo Garcia, Ivan Laptev, and Cordelia Schmid.
\newblock Segmenter: Transformer for semantic segmentation.
\newblock In \emph{ICCV}, pages 7242--7252, 2021.

\bibitem[Thawakar et~al.(2024)Thawakar, Narayan, Cholakkal, Anwer, Khan, Laaksonen, Shah, and Khan]{thawakar2024videoinstances}
Omkar Thawakar, Sanath Narayan, Hisham Cholakkal, Rao~Muhammad Anwer, Salman Khan, Jorma Laaksonen, Mubarak Shah, and Fahad~Shahbaz Khan.
\newblock Video instance segmentation in an open-world.
\newblock \emph{International Journal of Computer Vision}, 2024.

\bibitem[Toldo and Ozay(2022)]{9878745}
Marco Toldo and Mete Ozay.
\newblock Bring evanescent representations to life in lifelong class incremental learning.
\newblock In \emph{CVPR}, pages 16711--16720, 2022.

\bibitem[Wang et~al.(2021{\natexlab{a}})Wang, Wang, Shang-Guan, and Gupta]{9710595}
Jianren Wang, Xin Wang, Yue Shang-Guan, and Abhinav Gupta.
\newblock Wanderlust: Online continual object detection in the real world.
\newblock In \emph{ICCV}, pages 10809--10818, 2021{\natexlab{a}}.

\bibitem[Wang et~al.(2023)Wang, Xie, Zhang, Huang, Su, and Zhu]{wang2023hierarchical}
Liyuan Wang, Jingyi Xie, Xingxing Zhang, Mingyi Huang, Hang Su, and Jun Zhu.
\newblock Hierarchical decomposition of prompt-based continual learning: Rethinking obscured sub-optimality.
\newblock In \emph{NeurIPS}, 2023.

\bibitem[Wang et~al.(2024)Wang, Misra, Zeng, Girdhar, and Darrell]{Wang_2024_CVPR}
Xudong Wang, Ishan Misra, Ziyun Zeng, Rohit Girdhar, and Trevor Darrell.
\newblock Videocutler: Surprisingly simple unsupervised video instance segmentation.
\newblock In \emph{CVPR}, pages 22755--22764, 2024.

\bibitem[Wang et~al.(2021{\natexlab{b}})Wang, Xu, Wang, Shen, Cheng, Shen, and Xia]{9577282}
Yuqing Wang, Zhaoliang Xu, Xinlong Wang, Chunhua Shen, Baoshan Cheng, Hao Shen, and Huaxia Xia.
\newblock End-to-end video instance segmentation with transformers.
\newblock In \emph{CVPR}, pages 8737--8746, 2021{\natexlab{b}}.

\bibitem[Wang et~al.(2022)Wang, Zhang, Lee, Zhang, Sun, Ren, Su, Perot, Dy, and Pfister]{wang2022learning}
Zifeng Wang, Zizhao Zhang, Chen-Yu Lee, Han Zhang, Ruoxi Sun, Xiaoqi Ren, Guolong Su, Vincent Perot, Jennifer Dy, and Tomas Pfister.
\newblock Learning to prompt for continual learning.
\newblock In \emph{CVPR}, pages 139--149, 2022.

\bibitem[Wei et~al.(2024)Wei, Yang, Xu, and Deng]{wei2024class}
Kun Wei, Xu Yang, Zhe Xu, and Cheng Deng.
\newblock Class-incremental unsupervised domain adaptation via pseudo-label distillation.
\newblock \emph{IEEE Transactions on Image Processing}, 33:\penalty0 1188--1198, 2024.

\bibitem[Wei et~al.(2025)Wei, Xu, and Deng]{wei2025compress}
Kun Wei, Zhe Xu, and Cheng Deng.
\newblock Compress to one point: Neural collapse for pre-trained model-based class-incremental learning.
\newblock In \emph{AAAI}, pages 21465--21473, 2025.

\bibitem[Wu et~al.(2018)Wu, Herranz, Liu, wang, van~de Weijer, and Raducanu]{NIPS2018_7836}
Chenshen Wu, Luis Herranz, Xialei Liu, yaxing wang, Joost van~de Weijer, and Bogdan Raducanu.
\newblock Memory replay gans: Learning to generate new categories without forgetting.
\newblock In \emph{NeurIPS}, pages 5962--5972, 2018.

\bibitem[Wu et~al.(2022{\natexlab{a}})Wu, Jiang, Bai, Zhang, and Bai]{10.1007/978-3-031-19815-1_32}
Junfeng Wu, Yi Jiang, Song Bai, Wenqing Zhang, and Xiang Bai.
\newblock Seqformer: Sequential transformer for video instance segmentation.
\newblock In \emph{ECCV}, pages 553--569, 2022{\natexlab{a}}.

\bibitem[Wu et~al.(2022{\natexlab{b}})Wu, Liu, Jiang, Bai, Yuille, and Bai]{10.1007/978-3-031-19815-1_34}
Junfeng Wu, Qihao Liu, Yi Jiang, Song Bai, Alan Yuille, and Xiang Bai.
\newblock In defense of online models for video instance segmentation.
\newblock In \emph{ECCV}, pages 588--605, Cham, 2022{\natexlab{b}}.

\bibitem[Wu et~al.(2022{\natexlab{c}})Wu, Yarram, Liang, Lan, Yuan, Eledath, and Medioni]{9878426}
Jialian Wu, Sudhir Yarram, Hui Liang, Tian Lan, Junsong Yuan, Jayan Eledath, and Gérard Medioni.
\newblock Efficient video instance segmentation via tracklet query and proposal.
\newblock In \emph{CVPR}, pages 949--958, 2022{\natexlab{c}}.

\bibitem[Xie et~al.(2025)Xie, Lu, Xiao, Wang, Zhang, and Liu]{10.1007/978-3-031-73347-5_11}
Zhengyuan Xie, Haiquan Lu, Jia-wen Xiao, Enguang Wang, Le Zhang, and Xialei Liu.
\newblock Early preparation pays off: New classifier pre-tuning for class incremental semantic segmentation.
\newblock In \emph{ECCV}, pages 183--201, 2025.

\bibitem[Yang et~al.(2019)Yang, Fan, and Xu]{9008283}
Linjie Yang, Yuchen Fan, and Ning Xu.
\newblock Video instance segmentation.
\newblock In \emph{ICCV}, pages 5187--5196, 2019.

\bibitem[Yang et~al.(2021)Yang, Fang, Wang, Li, Fang, Shan, Feng, and Liu]{9710024}
Shusheng Yang, Yuxin Fang, Xinggang Wang, Yu Li, Chen Fang, Ying Shan, Bin Feng, and Wenyu Liu.
\newblock Crossover learning for fast online video instance segmentation.
\newblock In \emph{ICCV}, pages 8023--8032, 2021.

\bibitem[Ying et~al.(2023)Ying, Zhong, Mao, Wang, Chen, Wu, Liu, Fan, Zhuge, and Shen]{Ying_2023_ICCV}
Kaining Ying, Qing Zhong, Weian Mao, Zhenhua Wang, Hao Chen, Lin~Yuanbo Wu, Yifan Liu, Chengxiang Fan, Yunzhi Zhuge, and Chunhua Shen.
\newblock Ctvis: Consistent training for online video instance segmentation.
\newblock In \emph{ICCV}, pages 899--908, 2023.

\bibitem[Yoon et~al.(2018)Yoon, Yang, Lee, and Hwang]{yoon2018lifelong}
Jaehong Yoon, Eunho Yang, Jeongtae Lee, and Sung~Ju Hwang.
\newblock Lifelong learning with dynamically expandable networks.
\newblock In \emph{ICLR}, 2018.

\bibitem[Zenke et~al.(2017)Zenke, Poole, and Ganguli]{10555533058903306093}
Friedemann Zenke, Ben Poole, and Surya Ganguli.
\newblock Continual learning through synaptic intelligence.
\newblock In \emph{ICML}, page 3987–3995. JMLR.org, 2017.

\bibitem[Zhang et~al.(2024)Zhang, Yu, Dong, Li, Su, Chu, and Yu]{zhang2024mm}
Duzhen Zhang, Yahan Yu, Jiahua Dong, Chenxing Li, Dan Su, Chenhui Chu, and Dong Yu.
\newblock Mm-llms: Recent advances in multimodal large language models.
\newblock In \emph{ACL}, pages 12401--12430, 2024.

\bibitem[Zhang et~al.(2023)Zhang, Tian, Wu, Ji, Wang, Zhang, and Wan]{10376837}
Tao Zhang, Xingye Tian, Yu Wu, Shunping Ji, Xuebo Wang, Yuan Zhang, and Pengfei Wan.
\newblock Dvis: Decoupled video instance segmentation framework.
\newblock In \emph{ICCV}, pages 1282--1291, 2023.

\bibitem[Zhao et~al.(2021{\natexlab{a}})Zhao, Fu, Kang, Tian, Wu, and Li]{zhao2021mgsvf}
Hanbin Zhao, Yongjian Fu, Mintong Kang, Qi Tian, Fei Wu, and Xi Li.
\newblock Mgsvf: Multi-grained slow versus fast framework for few-shot class-incremental learning.
\newblock \emph{IEEE Transactions on Pattern Analysis and Machine Intelligence}, 46\penalty0 (3):\penalty0 1576--1588, 2021{\natexlab{a}}.

\bibitem[Zhao et~al.(2021{\natexlab{b}})Zhao, Wang, Fu, Wu, and Li]{zhao2021memory}
Hanbin Zhao, Hui Wang, Yongjian Fu, Fei Wu, and Xi Li.
\newblock Memory-efficient class-incremental learning for image classification.
\newblock \emph{IEEE Transactions on Neural Networks and Learning Systems}, 33\penalty0 (10):\penalty0 5966--5977, 2021{\natexlab{b}}.

\bibitem[Zhu et~al.(2024)Zhu, Liao, Zhang, Wang, Liu, and Wang]{vimicml2024}
Lianghui Zhu, Bencheng Liao, Qian Zhang, Xinlong Wang, Wenyu Liu, and Xinggang Wang.
\newblock Vision mamba: Efficient visual representation learning with bidirectional state space model.
\newblock In \emph{ICML}, 2024.

\bibitem[Ángeles Cerón et~al.(2022)Ángeles Cerón, Ruiz, Chang, and Ali]{CERON2022102569}
Juan~Carlos Ángeles Cerón, Gilberto~Ochoa Ruiz, Leonardo Chang, and Sharib Ali.
\newblock Real-time instance segmentation of surgical instruments using attention and multi-scale feature fusion.
\newblock \emph{Medical Image Analysis}, 81:\penalty0 102569, 2022.

\end{thebibliography}
}


\clearpage
\setcounter{page}{1}
\maketitlesupplementary

\appendix

\section{Experimental Settings}\label{sec: experimental_setting_supp}
\textbf{Benchmark Datasets:}
We conduct comparison experiments on three video instance segmentation datasets (\emph{i.e.}, YouTube-VIS 2019\footnote{https://codalab.lisn.upsaclay.fr/competitions/6064} \cite{9008283}, YouTube-VIS 2021\footnote{https://codalab.lisn.upsaclay.fr/competitions/7680} \cite{9008283} and OVIS\footnote{https://codalab.lisn.upsaclay.fr/competitions/4763} \cite{qi2021occluded}) to verify the effectiveness of our HVPL model in addressing the CVIS problem. 
\textbf{YouTube-VIS 2019} \cite{9008283} is the first dataset proposed to perform video instance segmentation and it has 40 semantic categories. we consider 20-2 and 20-5 settings on YouTube-VIS 2019. The 20-2 and 20-5 settings respectively indicate first learning 20 classes, followed by ten continual tasks, each with 2 new classes ($T=11$); and followed by 4 consecutive tasks, each with 5 new classes ($T=5$). 
\textbf{YouTube-VIS 2021} \cite{9008283} comprises the same semantic categories with YouTube-VIS 2019, but has more confusing trajectories than YouTube-VIS 2019. On YouTube-VIS 2021, we set 30-10 and 20-4 under the CVIS setting. The settings of 30-10 and 20-4 involve learning 30 classes followed by one task with 10 new classes ($T=2$), and learning 20 classes followed by 5 tasks with 4 new classes each ($T=6$). 
\textbf{OVIS} \cite{qi2021occluded}, which contains 25 classes, has distinct characteristics compared to YouTube-VIS 2021 \cite{9008283}: each video features more instances with heavy occlusions and diverse appearances. On OVIS \cite{qi2021occluded}, we set 15-5 and 15-10 for the CVIS problem. The settings of 15-5 and 15-10 involve learning 15 classes followed by 2 task with 5 new classes each ($T=3$), and learning 15 classes followed by one task with 10 new categories ($T=2$).

\begin{algorithm}[t]
\small
\caption{Optimization of The Proposed HVPL. }
\LinesNumbered
\label{alg: algorithm_pipeline}
\textbf{Initialize:} 
The pretrained Mask2Former and a sequence of video instance segmentation tasks $\mathcal{T}=\{\mathcal{T}^t\}_{t=1}^T$;  

\textcolor{blue}{$\triangleright$  \textbf{Training for The $t$-th Task: }} \\
Initialize the frame and video prompts $\{\mf{P}_{\mr{frm}}^t, \mf{P}_{\mr{vid}}^t\}$; \\
Initialize the classifier and mask heads $\{\Gamma_c^t, \Gamma_m^t\}$;\\

\For{($\mf{x}_i^t, \mf{y}_i^t) ~\mathrm{in} ~\mathcal{T}^t$}
{
\If{$t\geq 2$}
{
Update task-specific video prompt $\mf{P}_{\mr{vid}}^t$, classifier and mask heads $(\Gamma_c^t, \Gamma_m^t)$ via the loss in \cite{VITA}; \\

Compute the original gradient $\triangle\mf{P}$ used to update the task-specific frame prompt $\mf{P}_{\mr{frm}}^t$;\\

Conduct SVD on the feature space $\mathcal{O}^{t{-}1}$;\\

Obtain the orthogonal feature space $\widehat{\mf{V}}_0^{t{-}1}$;\\

Obtain $\triangle\mf{P}^*$ via gradient projection in Eq.~\eqref{eq: final_gradient};\\

Use $\triangle\mf{P}^*$ to update the frame prompt $\mf{P}_{\mr{frm}}^t$;
}

\Else{
Update learnable parameters via the loss in \cite{VITA};
}
}

Delete the feature space $\mathcal{O}^{t{-}1}$ of the $(t{-}1)$-th task;\\
Construct the feature space $\mathcal{O}^t$ for the $t$-th task;\\

\textbf{Return:} $\{\mf{P}_{\mr{frm}}^t, \mf{P}_{\mr{vid}}^t, \Gamma_c^t, \Gamma_m^t, \mathcal{O}^t \}$. \\

\textcolor{blue}{$\triangleright$ \textbf{Inference:}} \\
\textbf{Initialize:} All task-specific frame and video prompts $\{\mf{P}_{\mr{frm}}^t, \mf{P}_{\mr{vid}}^t\}_{t=1}^T$, along with the classifier and mask heads $\{\Gamma_c^t, \Gamma_m^t\}_{t=1}^T$, learned so far; \\

Obtain $\mf{P}_{\mr{frm}}^* = [\mf{P}_{\mr{frm}}^1; \mf{P}_{\mr{frm}}^2; \cdots; \mf{P}_{\mr{frm}}^T] \in\mathbb{R}^{TL_p^f\times D}$;\\

Obtain $\mf{P}_{\mr{vid}}^* = [\mf{P}_{\mr{vid}}^1; \mf{P}_{\mr{vid}}^2; \cdots; \mf{P}_{\mr{vid}}^T] \in\mathbb{R}^{TL_p^v\times D}$;\\

Obtain $\Gamma_c^*= [\Gamma_c^1, \Gamma_c^2, \cdots, \Gamma_c^T] \in\mathbb{R}^{D\times (|\mathcal{Y}^{1:t}| + 1)}$;\\

\textbf{Return:} Predictions about masks and class probabilities. 

\end{algorithm}

\begin{figure*}[t]
\centering
\includegraphics[trim = 0mm 22mm 0mm 22mm, clip, width=498pt, height=600pt]
{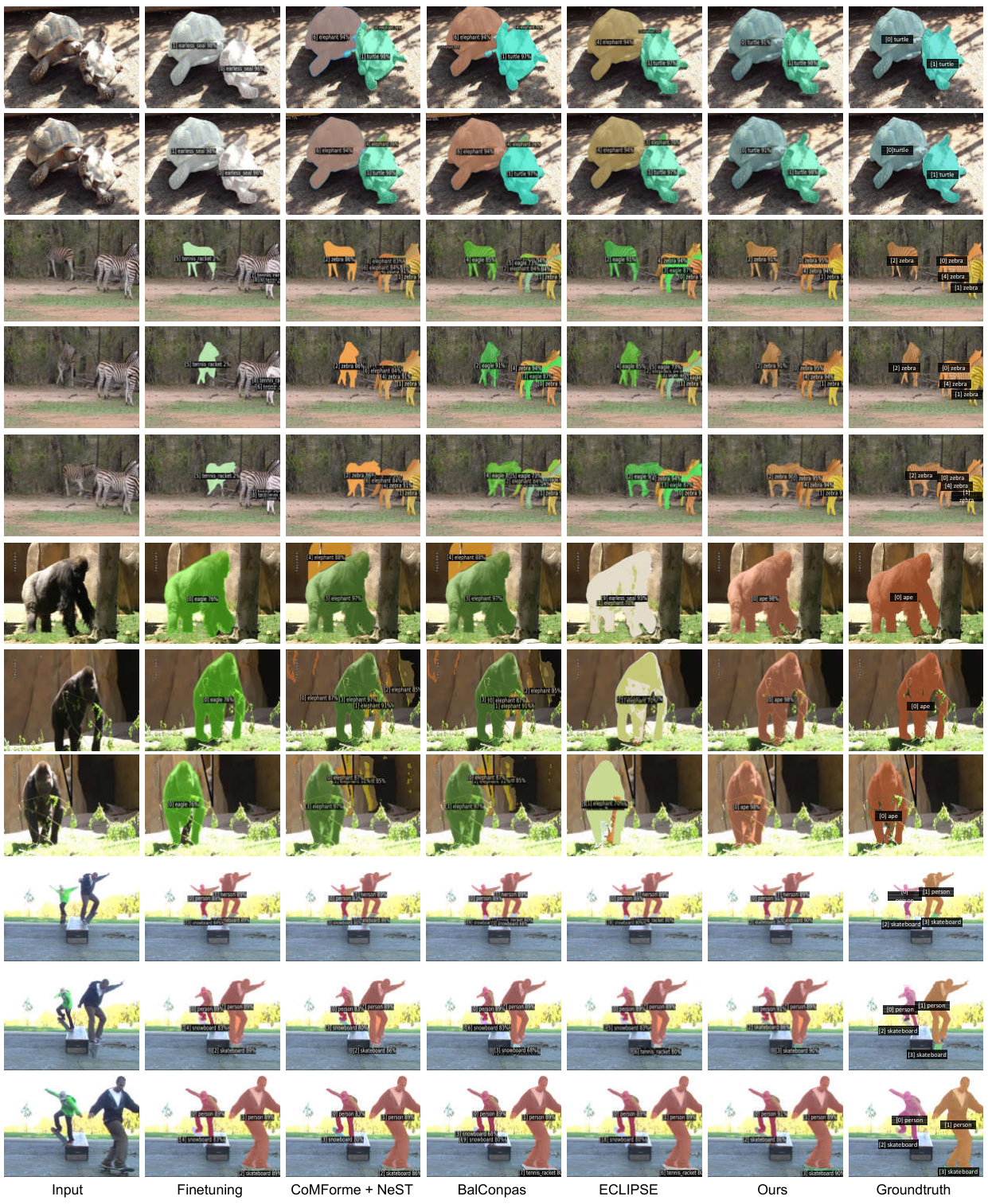}
\vspace{-20pt}
\caption{Comparison results of some selected video frames from YouTube-VIS 2019 \cite{9008283} under the 20-5 setting (zoom in for a better view). }
\label{fig: vis_YouTube2019_30_5_supp}
\end{figure*}

\begin{figure*}[t]
\centering
\includegraphics[trim = 26mm 0mm 26mm 0mm, clip, width=498pt, height=440pt]
{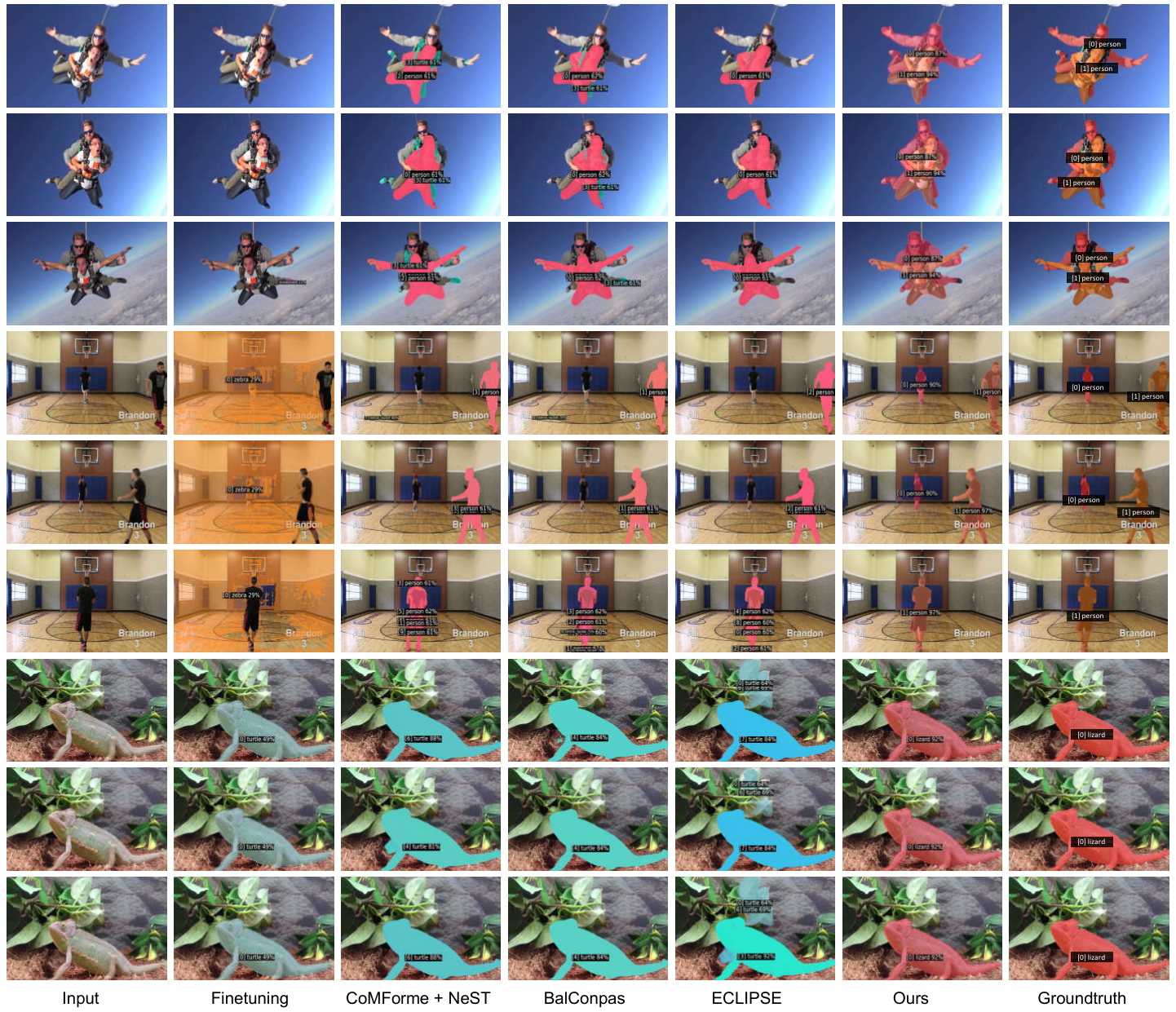}
\vspace{-22pt}
\caption{Comparisons of some selected video frames from YouTube-VIS 2021 \cite{9008283} under the 30-10 setting (zoom in for a better view). }
\label{fig: vis_YouTube2021_30_10_supp}
\vspace{-1mm}
\end{figure*}

\textbf{Implementation Details:}
For the network architecture, we introduce Mask2Former \cite{9878483} as the frame-level detector, where the backbone is ResNet-50 \cite{7780459}, the Transformer decoder $\mathcal{D}_{\mr{trans}}$ includes 9 MSA layers, and
the number of scales in the pixel decoder $\mathcal{D}_{\mr{pixel}}$ is 3. Besides, the video context decoder $\mathcal{D}_{\mr{video}}$ consists of 6 GSS layers and 3 MSA layers. All network parameters of the pretrained Mask2Former \cite{9878483} are frozen during the training phase.
Following \cite{dosovitskiy2021an}, we set the feature dimensions as $D=256$ and the number of heads in each MSA layer as 8. For the lengths of task-specific frame and video prompts, we set $L_p^f=100, L_p^v=100$ for the first VIS task and set $L_p^f=10, L_p^v=10$ to learn the $t$-th ($t\geq2$) incremental task. Furthermore, we adopt the same preprocessing strategy for input video frames during training and inference as proposed in \cite{VITA}. 
We optimize the parameters of the video context decoder $\mathcal{D}_{\mr{video}}$ at the first task, and freeze them when $t\geq 2$. If the learning of the $(t{-}1)$-th task is completed, we will store the feature space $\mathcal{O}^{t{-}1}$ in advance, and delete the feature space $\mathcal{O}^{t{-}2}$ learned at the $(t{-}2)$-th task for saving memory costs. Then $\mathcal{O}^{t{-}1}$ is utilized to learn the $t$-th task. 
To balance the forgetting of old tasks and the learning of new tasks, we empirically set $\xi=0.7$ to determine the orthogonal feature space $\widehat{\mf{V}}_0^{t{-}1}$ of the $(t{-}1)$-th task. Additionally, we utilize the Adam optimizer ($\beta_1=0.9, \beta_2=0.999$) to optimize the proposed HVPL model, where the initial learning rate is $5.0\times 10^{-5}$. 
The random seed is set to 42 in this paper.

\textbf{Comparative Methods:}
To exhibit the effectiveness of the proposed HVPL model, we introduce some state-of-the-art continual learning methods for comparison experiments. Specifically, MiB \cite{cermelli2020modeling} devises a novel objective function alongside a tailored classifier initialization strategy to address the issue of background shift. CoMFormer \cite{10203850} utilizes a new adaptive distillation loss combined with a mask-based pseudo-labeling technique to effectively mitigate forgetting. 
NeST \cite{10.1007/978-3-031-73347-5_11} proposes a classifier pretuning method, which is applied prior to the formal training process. Instead of directly adjusting the parameters of new classifiers, NeST learns a transformation from old classifiers to generate new classifiers for initialization. 
BalConpas \cite{chen2024strike} introduces a class-proportional memory strategy that ensures the class distribution in the replayed sample set aligns with the distribution in the historical training data.
ECLIPSE \cite{Kim_2024_CVPR} devises an efficient method for continual panoptic segmentation built on visual prompt tuning. For fair comparisons, all comparative methods employ the same backbone and data augmentation strategy for training.

\section{Optimization}\label{sec: optimization}
The optimization pipeline of our HVPL model is presented in \textbf{Algorithm}~\ref{alg: algorithm_pipeline}. During training, given a vide-label pair $(\mf{x}_i^t, \mf{y}_i^t)\in\mathcal{T}^t$ at the $t$-th ($t\geq 2$) task, 
we first update the task-specific video prompt $\mf{P}_{\mr{vid}}^t$, the classifier and mask heads $(\Gamma_c^t, \Gamma_m^t)$ via optimizing the loss proposed in \cite{VITA}. Then we compute the original gradient $\triangle\mf{P}$ used to update the task-specific frame prompt $\mf{P}_{\mr{frm}}^t$, and conduct SVD on the feature space $\mathcal{O}^{t{-}1}$ of the $(t{-}1)$-th task. After obtaining the orthogonal feature space $\widehat{\mf{V}}_0^{t{-}1}$, we perform gradient projection to derive $\triangle\mf{P}^* = \triangle\mf{P}\widehat{\mf{V}}_0^{t{-}1}(\widehat{\mf{V}}_0^{t{-}1})^\top$ via Eq.~\eqref{eq: final_gradient}, and employ $\triangle\mf{P}^*$ to update the task-specific frame prompt $\mf{P}_{\mr{frm}}^t$. During inference, we first concatenate all frame prompts $\{\mf{P}_{\mr{frm}}^t\}_{t=1}^T$ as $\mf{P}_{\mr{frm}}^* = [\mf{P}_{\mr{frm}}^1; \mf{P}_{\mr{frm}}^2; \cdots; \mf{P}_{\mr{frm}}^T] \in\mathbb{R}^{TL_p^f\times D}$, all task-specific video prompts $\{\mf{P}_{\mr{vid}}^t\}_{t=1}^T$ as $\mf{P}_{\mr{vid}}^* = [\mf{P}_{\mr{vid}}^1; \mf{P}_{\mr{vid}}^2; \cdots; \mf{P}_{\mr{vid}}^T] \in\mathbb{R}^{TL_p^v\times D}$, and all classifier heads $\{\Gamma_c^t\}_{t=1}^T$ as $\Gamma_c^*= [\Gamma_c^1, \Gamma_c^2, \cdots, \Gamma_c^T] \in\mathbb{R}^{D\times (|\mathcal{Y}^{1:t}| + 1)}$. Subsequently, we utilize them to predict masks and class probabilities for a given test video.

\section{Qualitative Comparisons}\label{sec: qualitative_comparisons}
As shown in Figs.~\ref{fig: vis_YouTube2019_30_5_supp}--\ref{fig: vis_YouTube2021_30_10_supp}, to evaluate the performance of our proposed model in various CVIS scenarios, we present the visualization results of selected video frames from  YouTube-VIS 2019 and YouTube-VIS 2021 \cite{9008283}. The following observations can be drawn from the results: 1)
Significant Improvement Over Prompt Learning-Based Methods: The proposed model outperforms the prompt learning-based method ECLIPSE \cite{Kim_2024_CVPR} across all settings, demonstrating superior capability in addressing CVIS problem. Notably, in complex backgrounds and dynamic scenarios, our model mitigates catastrophic forgetting at both the frame and video levels. 2) Better Performance Than Knowledge Distillation-Based Methods: Compared to knowledge distillation methods such as CoMFormer \cite{10203850}, NeST \cite{10.1007/978-3-031-73347-5_11} and BalConpas \cite{chen2024strike}, our model achieves more precise instance segmentation, especially in cases involving multiple or small objects. This highlights the effectiveness of the task-specific video prompt and the video context decoder in capturing global video contexts to tackle catastrophic forgetting. These visual results further validate the effectiveness of the proposed model in tackling forgetting of old classes at both the frame and video levels.

\section{Societal Impact and Limitations} 
\label{sec: impact_and_limitations}
\textbf{Societal Impact:}
Continual Video Instance Segmentation (CVIS) is an emerging research field at the intersection of computer vision and continual learning, focusing on the ability to segment, track, and incrementally learn from objects in video streams over time. Unlike traditional video instance segmentation, which operates on fixed, predefined datasets, CVIS emphasizes continual learning, enabling models to incorporate new information dynamically without retraining from scratch. Continual Video Instance Segmentation (CVIS) has the potential to significantly influence various aspects of society by enabling machines to dynamically learn and adapt from video data over time. 
\begin{itemize}
\item Improved Automation and Efficiency: CVIS enhances automation in applications like autonomous vehicles, smart cities, and industrial monitoring, enabling real-time adaptation to changing environments and improving efficiency.

\item Enhanced Public Safety: By enabling better anomaly detection and situational awareness in video surveillance and management, CVIS contributes to safer communities. 

\item Environmental and Wildlife Monitoring: CVIS can support long-term ecological studies, enabling the tracking of wildlife and monitoring of environmental changes without continuous human intervention. 

\end{itemize}

\textbf{Limitations:}
Although the proposed HVPL model can address the CIVS problem by alleviating catastrophic forgetting from both the image-level and video-level perspectives, it may struggle to continually learn large-scale semantically similar tasks. Therefore, we will explore how to increase the scalability of our proposed HVPL model in the future.

\end{document}